%% file: webmre.tex
\documentclass[11pt]{article}
\usepackage{acl}
\usepackage{times}
\usepackage{latexsym}
\usepackage[T1]{fontenc}
\usepackage[utf8]{inputenc}
\usepackage{microtype}
\usepackage{inconsolata}
\usepackage{graphicx}
\usepackage{amsmath}
\usepackage{amssymb}
\usepackage{multicol}
\usepackage{booktabs}
\usepackage[most]{tcolorbox}

\newtcolorbox{promptbox}[1]{enhanced, colback=black!4, colframe=black!75, colbacktitle=black!75, coltitle=white, fonttitle=\bfseries\small, title={#1}, boxrule=0.5pt, arc=1.5pt, left=5pt, right=5pt, top=5pt, bottom=5pt}

\title{Guides That Cause Actions: An Offline Study of Guide-Action Mutual Reinforcement in Multimodal Web Agents}

\author{
  Chengguang Gan\textsuperscript{1} \quad Yunhao Liang\textsuperscript{2} \quad Qinghao Zhang\textsuperscript{3} \quad Shiwen Ni\textsuperscript{4} \\[3pt]
  \textsuperscript{1}Independent Researcher \quad \textsuperscript{2}University of Chinese Academy of Sciences \\
  \textsuperscript{3}Pusan National University \quad \textsuperscript{4}Shenzhen University of Advanced Technology \\[3pt]
  \normalsize Correspondence: \texttt{chengguangg1024@gmail.com}
}

\begin{document}
\maketitle

\begin{abstract}
Web agents are usually evaluated in live environments, where environment state and judge models drift between runs, so the same checkpoint rarely reproduces the same score, making controlled studies of training phenomena impractical. We present WebMRE, an offline benchmark of 541 tasks and 5,293 steps derived from successful WebArena trajectories, with fully audited test labels and a deterministic protocol that scores a checkpoint identically on every run without any environment. Each step pairs a human oriented guide sentence with a grounded action, enabling the first study of the mutual reinforcement effect between them in web agents. Averaged over three seeds the effect holds for both models in both decoding orders and grows with scale: jointly decoding a guide lifts element selection over an action only reference by 0.9 and 0.2 points for Qwen3.5-4B and by 1.7 and 2.2 points for Qwen3.5-9B. A mediation analysis shows that the guide is a causal channel rather than commentary: forcing the gold guide as a decoding prefix lifts action accuracy from .422 to .684, another step's guide collapses it to .055, and a paraphrase that renames the target still recovers half of the gain, so the channel carries instruction meaning and not only the label string. The same channel yields an offline reward that only a replayable protocol makes computable, though optimizing it from a strong checkpoint brings no gain yet. Our fine tuned models outperform GPT-5.5, Claude Opus 4.8, and Gemini 3.5 Flash, run zero shot, on every offline metric.
\end{abstract}

\begin{figure}[t]
  \centering
  \includegraphics[width=\columnwidth]{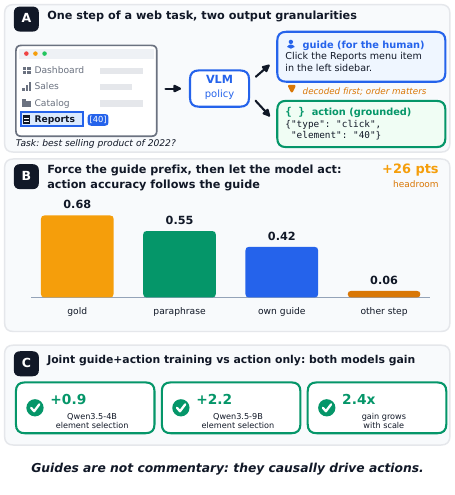}
  \caption{The web guide agent task (A) and the two findings of this paper (B, C), walked through in Section~\ref{sec:intro}.}
  \label{fig:teaser}
\end{figure}

\section{Introduction}
\label{sec:intro}

Multimodal web agents are almost always evaluated by executing tasks in live sandboxes such as WebArena and its visual extension \citep{zhou2024webarena,koh2024visualwebarena}, and the field largely equates progress with live success rate \citep{nguyen2025gui}. A live protocol is the right instrument for measuring end to end competence, but it is a poor instrument for studying how an agent should be trained. Authentication state expires, backend content drifts, and judge models change their verdicts between runs; in our preliminary live evaluations, the same proprietary agent scored up to twenty points apart across two deployments of the same benchmark. When single digit differences decide a conclusion, this noise floor rules out controlled comparisons between training choices.

This paper studies one such training question on a deliberately offline footing. The task, shown in Figure~\ref{fig:teaser}A, comes from digital adoption practice \citep{gan2026mag}: at every step of a web task, the model receives a screenshot, the task intent, and the interaction history, and produces two outputs of different granularity. The coarse output is a guide, one sentence that tells a human user what to do on the current screen. The fine output is the action itself, grounded either in a marked element or in pixel coordinates. We build WebMRE, a corpus of 541 tasks and 5,293 steps derived from successful WebArena trajectories, re-audit every flagged test label with a panel of three judge models, and score checkpoints with a deterministic protocol: greedy decoding and rule based matching, so the same checkpoint yields the same numbers on every run, with no environment in the loop.

The question is whether the two granularities reinforce each other. Prior work on the mutual reinforcement effect found that coarse and fine information extraction tasks improve each other when trained jointly \citep{gan-etal-2026-multilingual}, and GUI training pipelines routinely add instruction like text to action data on the assumption that more supervision helps. Our answer, measured on the natively multimodal Qwen3.5 family across three training seeds, is that the effect holds in web agents and grows with model scale (Figure~\ref{fig:teaser}C). Joint guide and action training improves element selection over an action only reference for both models and in both decoding orders: the action first order adds 0.9 points at 4B and 1.7 at 9B, the guide first order 0.2 and 2.2. The coarse guide and the fine action therefore reinforce each other, and a larger model with stronger priors extracts more from the shared supervision, which is what the mutual reinforcement account predicts.

Why does the guide carry this influence? We intervene directly on the causal path (Figure~\ref{fig:teaser}B): force a guide prefix into the decoder, then let the model complete the action. With its own guide the model scores as in free decoding, a check that the intervention itself is neutral. With the gold guide, action accuracy rises to .684; with another step's guide it collapses to .055. The model does not merely emit guides, it follows them. A paraphrase control, which renames the target by function rather than by its visible label, still recovers half of the gold gain, so the channel carries instruction meaning and not only surface overlap. The guide is a causal channel into the action, and the quality of generated guides, which a blind panel rates useful only half as often as gold ones, is the current bottleneck.

A causal channel can also be paid for causally. Mediated-Guide GRPO turns the intervention above into a reward: a sampled guide earns credit according to whether the action it induces is correct, rather than how closely it matches a reference. Such a reward is computable only offline, because it re-queries the policy under a forced prefix on a frozen observation, and it needs no judge model. In a small two round budget it ends above an action only control but does not improve on an already strong supervised checkpoint, so we report the reward construction and its negative optimization result in Appendix~\ref{app:grpo}. As a byproduct of the same protocol, our fine tuned models outperform zero shot GPT-5.5, Claude Opus 4.8, and Gemini 3.5 Flash on every offline metric, including guide usefulness, an expected in distribution result that also validates the scorer.

Our contributions are:
\begin{itemize}
  \item WebMRE, an audited and view unified re-release of the MAG trajectory corpus together with a deterministic offline evaluation protocol for the two granularity web guide agent task; our additions are a three judge label audit (225 steps relabelled, 128 marked unmappable), 859 restored coordinate only steps, and per step staleness flags;
  \item the first measurement of the mutual reinforcement effect in web agents, showing that joint guide and action training improves element selection for both Qwen3.5-4B and Qwen3.5-9B, in both decoding orders, with the gain growing with model scale;
  \item causal evidence, from forced prefix interventions with a paraphrase control, that guides drive actions;
  \item Mediated-Guide GRPO, an offline reward that pays a guide by the action it causes, which the deterministic protocol makes computable without any environment, together with a negative result on optimizing it from an already strong supervised checkpoint.
\end{itemize}

\begin{figure*}[t]
  \centering
  \includegraphics[width=\textwidth]{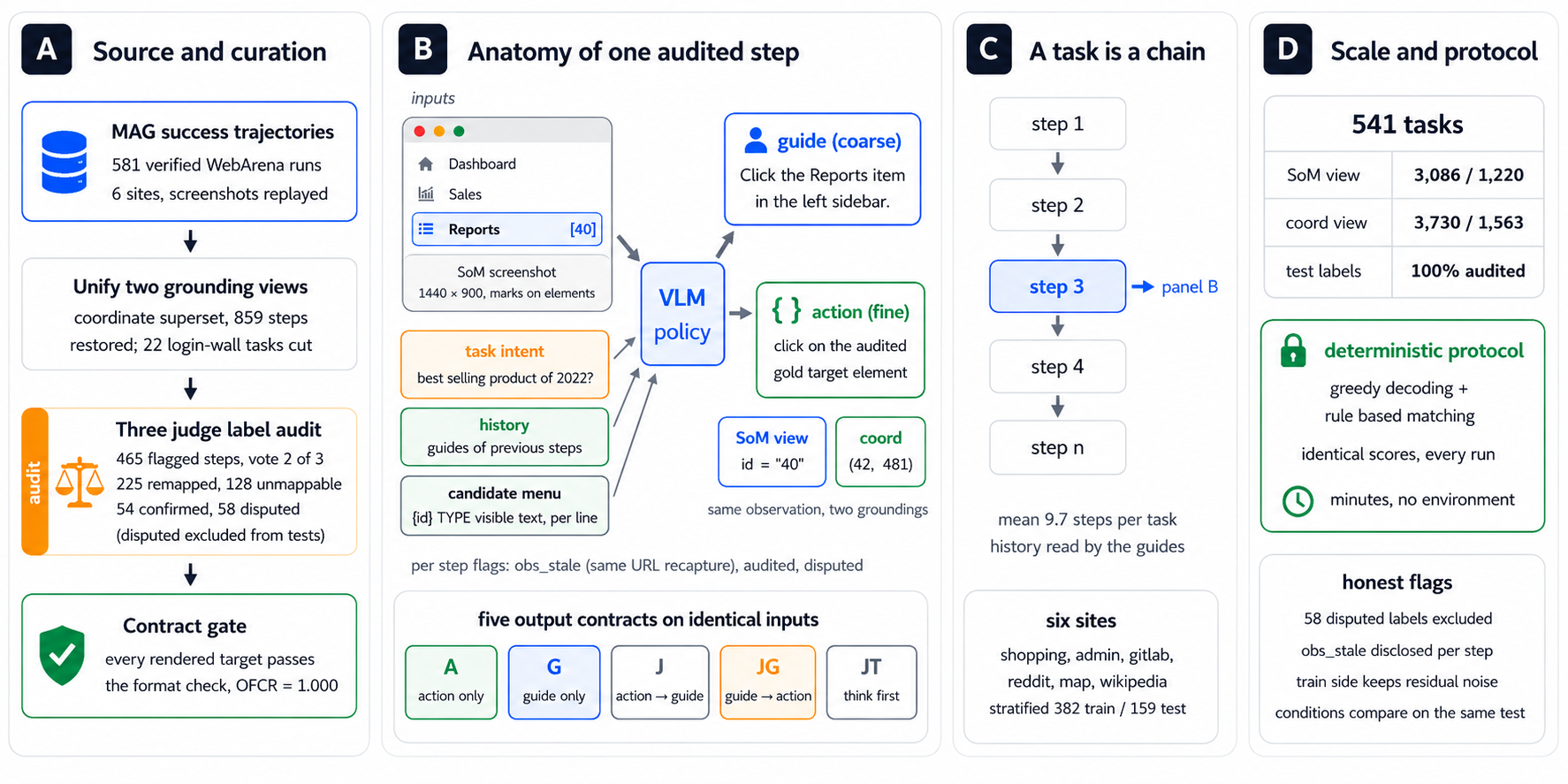}
  \caption{The WebMRE dataset. (A)~Curation from MAG trajectories, with a three judge label audit. (B)~One step: four inputs, two output granularities, two groundings of the same observation, and the output contracts rendered from identical inputs. (C)~A task is a chain of steps whose guides thread into later inputs. (D)~Scale and the deterministic offline protocol.}
  \label{fig:dataset}
\end{figure*}

\section{Related Work}
\label{sec:related}

Web agent evaluation is dominated by live benchmarks. WebArena and VisualWebArena execute agents against hosted websites and score functional task success \citep{zhou2024webarena,koh2024visualwebarena}, and MAG extends this setting with per step guide annotations and a live harness \citep{gan2026mag}. Offline corpora take the opposite trade: Mind2Web and WebLINX score recorded steps without an environment \citep{deng2023mind2web,lu2024weblinx}, and AgentTrek synthesizes trajectories from web tutorials \citep{xu2024agenttrek}. These corpora supply actions at scale, but none pairs every action with a guide written for a human user, and none treats the reliability of its own labels as an object of study; WebMRE provides both.

A parallel line augments GUI action data with text. Aguvis attaches low level instructions and inner monologue to grounding data \citep{xu2024aguvis}, Android in the Zoo interleaves chains of action thought \citep{zhang2024android}, and UI-TARS trains on thought traces at scale \citep{qin2025ui}. In all of these the text is auxiliary supervision assumed to help. Whether it helps, in which direction, and under which decoding order is rarely measured across seeds; our four contract comparison quantifies exactly this, and separates the effects that survive seed variation from those that do not.

Reinforcement learning for web and GUI agents optimizes the action against environment feedback or rule based rewards. WebRL and WebAgent-R1 train in live environments on task success \citep{qi2025webrl,wei2025webagent}, while UI-R1 and GUI-R1 apply group relative policy optimization with verifiable action rewards \citep{lu2026ui,luo2025gui}, building on GRPO \citep{shao2024deepseekmath} and its refinements \citep{liu2025understanding,NEURIPS2025_a4277440}. In every case the reward attaches to the action or the trajectory; the textual output, when present, is scored by similarity or ignored. Mediated-Guide GRPO instead prices the guide by the action it causes, with no environment and no learned judge in the loop.

This paper differs from all three lines in its object of study. We build an offline web agent dataset whose evaluation is free of environment variance, reproduces exactly across runs, and scores a checkpoint in under one GPU hour, and we import the mutual reinforcement effect, previously established in information extraction \citep{gan-etal-2026-multilingual}, into the web guide agent task. The corpus derives from MAG trajectories, but where MAG contributes a live benchmark and harness, our contribution is the phenomenon and its mechanism: an empirical demonstration that guide and action supervision reinforce each other conditionally, and an interpretability account of why.

\section{The WebMRE Dataset and Offline Protocol}
\label{sec:dataset}

\subsection{Provenance and Unification}
\label{sec:provenance}

WebMRE starts from the corpus released with MAG \citep{gan2026mag}: 581 WebArena trajectories that a live evaluator verified as successful, covering 563 distinct tasks across six sites, with each step replayed into a Set-of-Marks (SoM) screenshot, a gold action, and a guide sentence. MAG ships these steps in two parallel views, one grounding the action in a marked element and one in pixel coordinates, and the two views cover different subsets of steps. We unify them into a single corpus keyed by task and step index. The coordinate view is the superset; the element id is kept where the target was marked and left empty where it was not; and 859 steps that previously survived only in the coordinate view return to the corpus, restoring guide chains that the element view had truncated. Twenty two tasks whose screenshots were recaptured behind a login wall are removed. Because screenshots come from revisiting each step's URL, a consecutive step on the same URL may not reflect state changes inside the page; we flag such recaptures per step (obs\_stale, 59 percent of steps) rather than hide them, and Section~\ref{sec:mre} reports scores split by the flag. The result is 541 tasks with 5,293 steps (Figure~\ref{fig:dataset}); Figure~\ref{fig:dataexample} shows two records.

\input{tables/box_dataexample}

\subsection{Three Judge Label Audit}
\label{sec:audit}

The element labels in the source views come from an automatic mapping of recorded click coordinates onto marked boxes, and this mapping is a known noise source. We recall suspects with two heuristics. The first compares token F1 between the guide and the mapped element's text against the best scoring alternative, and flags 100 steps. It misses short labels: on the corpus' very first step the guide says to click the Reports menu item, the recorded target reads Pending Reviews, and the true element, labeled just Reports, scores 0.13 against a fourteen word guide. The second heuristic therefore flags any click whose recorded target shares no content word with its guide while another candidate's text is contained in the guide. Together the two rounds flag 1,108 of roughly 2,500 clicks; we audit every flagged test step (365) plus the full first round pool (100), and leave flagged training steps unaudited, a residual noise that every training condition shares equally.

Each audited step goes to three judge models from three vendors, which see the screenshot, the guide, the recorded target, the click coordinates, and the full candidate menu, and independently return confirm, remap with an id, or unmappable. A verdict applies only with a two of three majority; a remap additionally requires two judges to name the same id; steps without a majority keep their original label, are flagged, and are excluded from element level scoring. Of the 465 audited steps, 225 are remapped, 128 are unmappable and become coordinate only, 54 are confirmed, and 58 are disputed; 51 of those disputed steps fall in the SoM test split and are the ones excluded from element level scoring throughout. On the first step example above, all three judges name Reports, and the three proprietary baselines of Section~\ref{sec:mre}, run zero shot, later select the same element. One judge model also serves as a baseline later; remaps require agreement across vendors, and that model ends up the weakest baseline, so the overlap buys it no advantage. The same blind panel protocol governs guide text: guides flagged by an anchor sweep are rewritten by one model and accepted only when the panel rates the rewrite useful at least twice and no worse than the original, which replaces 18 of 100 candidates.

\begin{table}[t]
\centering
\small
\begin{tabular}{lrr}
\toprule
 & SoM view & Coordinate view \\
\midrule
Train steps & 3,086 & 3,730 \\
Test steps & 1,220 & 1,563 \\
\addlinespace[2pt]
Tasks (train / test) & \multicolumn{2}{c}{541 \;(382 / 159)} \\
Sites & \multicolumn{2}{c}{6} \\
Audited test labels & \multicolumn{2}{c}{100\% of flagged} \\
Steps flagged obs\_stale & \multicolumn{2}{c}{59\%} \\
\bottomrule
\end{tabular}
\caption{WebMRE statistics. The two views share tasks, splits, screenshots, and guides; they differ in the action grounding and in which steps admit an element id.}
\label{tab:stats}
\end{table}

\subsection{Task Suite and Offline Metrics}
\label{sec:metrics}

Every experimental condition sees the same inputs: the SoM screenshot, the task intent, the guides of previous steps, and the candidate menu. What varies is only the output contract, a fixed key set the model must emit as one JSON object. We train four contracts: action only (A), guide only (G), action then guide (J), and guide then action (JG). The corpus ships two grounding views that share the observation and differ solely in the action argument, an element id in the SoM view and an $(x, y)$ pixel position in the coordinate view; the experiments in this paper use the SoM view. Table~\ref{tab:stats} summarizes the corpus.

Let $\mathcal{D}$ be the test steps of a view, excluding disputed labels for element level terms. Each prediction is scored by a format gate and three matchers:
\begin{equation}
\mathrm{SAA} \;=\; \frac{1}{|\mathcal{D}|}\sum_{i\in\mathcal{D}} F_i \, M^{\mathrm{type}}_i \, M^{\mathrm{tgt}}_i \, M^{\mathrm{cnt}}_i .
\label{eq:saa}
\end{equation}
The gate $F_i\!\in\!\{0,1\}$ requires a parseable object with exactly the contract's keys, legal action arguments, and a guide free of ids and coordinates; its mean over $\mathcal{D}$ is reported as OFCR. $M^{\mathrm{type}}_i$ is an exact match on the action type. The target matcher depends on the view:
\begin{equation}
M^{\mathrm{tgt}}_i =
\begin{cases}
\mathbf{1}\,[\hat{e}_i = e_i] & \text{SoM view},\\[2pt]
1 & \hat{p}_i \in B_i,\\[2pt]
\exp\!\big({-d(\hat{p}_i, B_i)}/{\tau}\big) & \text{otherwise},
\end{cases}
\label{eq:tgt}
\end{equation}
where $e_i$ is the audited gold id, $B_i$ the gold element's box, $d$ the distance from the predicted point $\hat{p}_i$ to the box edge, and $\tau=120$ pixels; actions without a target require an empty argument. The content matcher gates typed text through token F1,
\begin{equation}
M^{\mathrm{cnt}}_i = \mathbf{1}\!\left[\mathrm{F1}(\hat{c}_i, c_i)\ge\theta_a\right],
\label{eq:cnt}
\end{equation}
with $\theta_a = 0.9$ for type and select, $0.5$ for finish, and $0$ otherwise.
Guides are scored against the gold sentence with BLEU \citep{papineni2002bleu} and ROUGE \citep{lin2004rouge}, and by usefulness: three judge models blind rate a guide as useful, ambiguous, or useless given the screenshot and the gold action, and we report the majority share of useful.

Every quantity above is a pure function of the model's greedy output, so a checkpoint maps to exactly one score vector: two evaluations of the same model cannot disagree. A full test pass costs under one GPU hour and scoring is immediate; no environment, no retries, and no judge sit in the action metrics.

\begin{figure*}[t]
  \centering
  \includegraphics[width=\textwidth]{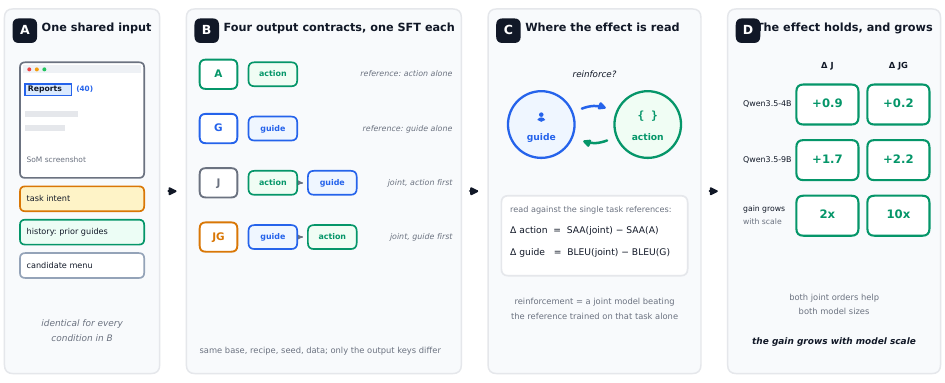}
  \caption{Measuring the mutual reinforcement effect. (A)~All conditions share identical inputs. (B)~The output contracts, one model each; A and G are the single task references. (C)~The effect is read as the joint model's gain over the reference of each granularity. (D)~Preview of Table~\ref{tab:multiseed}: both joint orders help both models, and the gain grows with scale.}
  \label{fig:mre}
\end{figure*}

\input{tables/table2_main}

\section{Mutual Reinforcement Effect in the Web Guide Agent Task}
\label{sec:mre}

\subsection{Setup}
Figure~\ref{fig:mre} lays out the design. Every condition fine tunes all parameters of Qwen3.5-4B \citep{qwen35blog} for three epochs with learning rate $10^{-5}$, cosine schedule, and effective batch eight, over three seeds; only the output contract changes. The same contracts are repeated on Qwen3.5-9B \citep{qwen35blog}, the larger model in the same natively multimodal family, under optimizer state sharding. Four output contracts are trained: action only (A), guide only (G), and the two joint orders J (action then guide) and JG (guide then action). GPT-5.5, Claude Opus 4.8, and Gemini 3.5 Flash run zero shot under contract J with the same prompts, images, and scorer.

\subsection{Results}
\label{sec:mainresults}
Table~\ref{tab:main} holds the grid; three findings matter.

\paragraph{The mutual reinforcement effect holds and grows with scale.} Decoding a guide together with the action improves element selection over the action only reference at both model sizes (Table~\ref{tab:multiseed}, means over three seeds). The action first order (J) adds $+0.9$ points at 4B and $+1.7$ at 9B; the guide first order (JG) adds $+0.2$ and $+2.2$. Both joint orders help both models, so the coarse guide and the fine action reinforce each other, and the gain grows with model scale, from under a point at 4B to more than two points at 9B. The 9B backbone starts from a weaker action only reference (.404 against .423), so the largest absolute score in the grid is the 4B action first model at .432; what scale changes is how much the joint contracts add on top of their own reference.

\paragraph{The guide direction is weaker, and order dependent.} On the guide side the joint contracts stay close to the guide only reference rather than beating it: at 9B, BLEU-1 is .551 for J and .558 for JG against .563 for G, and ROUGE-L .568 and .578 against .582. The action side therefore carries the effect. Decoding order matters here in the opposite way from the action side: at 9B the guide first order writes the better guide (.558 versus .551 BLEU-1) and also gains the most on the action, while at 4B the action first order is better on both (.551 versus .545). A guide decoded after the action is conditioned on a committed decision and can only describe it, whereas a guide decoded first must be produced from the observation alone, which is harder but is what makes it useful to the action that follows. Section~\ref{sec:channel} tests that reading directly.

\paragraph{Format compliance is not the differentiator.} All fine tuned rows emit parseable objects almost always: OFCR is .999 for the action only contract at both sizes and .984 to .991 for the joint contracts, so the joint models pay a fraction of a point for the longer output and nothing more. Because SAA is gated by this format check (Equation~\ref{eq:saa}), a contract cannot win by being easier to parse; the differences in Table~\ref{tab:main} are differences in the actions themselves.

\paragraph{Fine tuning clears the zero shot frontier.} The best fine tuned row beats the best zero shot row by 17 action points and 20 BLEU-1 points on guides. The three frontier models land close together on the action, .259 for GPT-5.5, .261 for Claude Opus 4.8 and .204 for Gemini 3.5 Flash, and in the same order on the guide, .359, .363 and .238 BLEU-1. Two things separate them from the fine tuned rows. The first is grounding: GPT-5.5 and Claude parse cleanly (OFCR .989 and .967), so almost every one of their outputs is scored and their gap is a gap in choosing the right element, not in producing valid JSON. The second is format: Gemini emits a parseable object on only .705 of steps, and since SAA is gated by that check its action score is depressed by roughly the same margin, which reproduces offline the model specific behavior that MAG reported in a live harness \citep{gan2026mag}. This is the expected outcome of task specific training rather than a capability claim over much larger models, and it mainly serves to show that the scorer separates models the way a live evaluation does.

\subsection{Robustness}
\label{sec:slices}
Five of the six model and contract combinations are positive on all three seeds; the exception is 4B guide first, negative on one seed by 0.3 points. Per seed values are in Table~\ref{tab:seeds}.

\input{tables/table_multiseed}

\paragraph{Sites.} The gain is not carried by one site. Pooling the three seeds, a joint contract is the best of the three rows on five of the six sites for each model (Table~\ref{tab:fullsite}); the one site where the action only reference wins is reddit at 4B and shopping at 9B. The two hardest and the two easiest sites move together: the administrative shop, the hardest site for element selection, improves from .262 to .292 at 9B, and gitlab, the largest site in the test split, from .482 to .523. Where a joint contract does lose to the action only reference, the loss is confined to a single site rather than spread across the corpus.

\input{tables/appendix_fullresults}

\paragraph{Action types.} The gain concentrates where the step requires locating something on the screen. Pooling the three seeds at 9B, the best joint contract adds 3.3 points on clicks (.398 to .431 over 599 steps) and 2.2 points on typing (.361 to .383 over 282 steps), the two types that must identify a target element, while finish steps do not move at all (.363 for all three contracts) and scrolling gains 1.8 points. At 4B the same pattern holds with typing gaining most (.374 to .409). Finish steps are the informative null: they carry no target and only ask for the final answer string, so a guide that describes what to do on the screen has nothing to contribute, and indeed contributes nothing. The mutual reinforcement effect thus acts through the part of the action that needs grounding, which is what the causal analysis in Section~\ref{sec:channel} then examines directly.

\paragraph{Staleness.} Because 59 percent of steps reuse the screenshot of the preceding step (Section~\ref{sec:provenance}), we score the flagged and the unflagged slice separately (Table~\ref{tab:fullstale}). Both joint contracts gain on the flagged slice at both model sizes, and on the unflagged slice they stay at or above the action only reference except for 4B guide first (.459 against .472), so the effect does not come from the recapture disclosure. Flagged steps are uniformly harder for every contract, which is what a recaptured screenshot predicts: at 9B the action only reference scores .372 on flagged steps against .472 on unflagged ones. That is also where the joint contracts help most, since the guide of the previous step carries what the stale screenshot does not show.

\paragraph{Templates.} Fifteen of the 159 test tasks share no intent template with any training task, and the joint gain persists on this slice, so the model is not merely reproducing memorized guide and action pairs.

\section{Analysis of Guides as a Causal Channel}
\label{sec:channel}

\begin{figure*}[t]
  \centering
  \includegraphics[width=\textwidth]{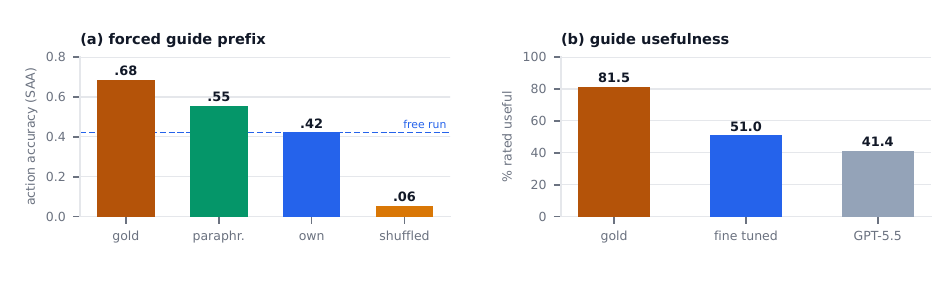}
  \caption{The guide is a causal channel. (a)~Action accuracy of the Qwen3.5-9B guide first model under forced guide prefixes: the gold guide lifts accuracy far above free decoding, a paraphrase that renames the target recovers half of the gain, and a guide belonging to another step collapses it. (b)~Blind three judge usefulness of gold and generated guides.}
  \label{fig:mechanism}
\end{figure*}

\paragraph{Inference time mediation.} On all 1,220 SoM test steps we force the decoder to begin its output with a chosen guide and let it complete only the action (Figure~\ref{fig:mechanism}a; a worked example in Appendix~\ref{app:examples}). Forcing the 9B model's own guide reproduces free decoding (.422), so the intervention itself is neutral. The gold guide lifts action accuracy to \textbf{.684}, and a guide taken from a different step collapses it to \textbf{.055}, a gold minus shuffled gap of 62.9 points on the same steps and the same checkpoint. The model does not merely emit a guide next to the action, it executes the guide it is given, and the .68 ceiling under gold guides says the binding constraint is guide quality rather than grounding ability. A leakage control rules out string copying: rewriting each gold guide so that it names the target by function rather than by its visible label still reaches .554, half of the .262 gain that the gold guide adds over free decoding, so half of the channel survives with the target renamed and the guide carries instruction meaning and not only the label string. Table~\ref{tab:channel} collects the five conditions.

\input{tables/table_channel}

\paragraph{Usefulness.} Three judge models blind rate guides given the screenshot and the gold action (Figure~\ref{fig:mechanism}b). Gold guides rate 81.5 percent useful, matching the 82 percent human study on the source corpus \citep{gan2026mag}, which calibrates the panel; the judges agree pairwise on .93 to .97 of items, so no single judge drives the rating. Generated guides rate 51.0 percent for the fine tuned model and 41.4 percent for GPT-5.5: the fine tuned model wins the usefulness comparison too, but half of its guides still fail, the guide and the action wrong together as the mediation coupling predicts (Appendix~\ref{app:examples}).

\paragraph{Guide quality is the binding constraint.} The two measurements above bound the headroom of this task from both sides. A model that writes gold quality guides would reach .684 on the action, 26 points above what the same checkpoint achieves when it writes its own guide, and the paraphrase condition shows that this headroom is about instruction content rather than surface form. Yet only half of the generated guides are rated useful, so the gap between .422 and .684 is largely a guide generation gap, not a grounding gap. This reframes what the mutual reinforcement effect buys: the joint contracts of Section~\ref{sec:mainresults} improve the action by one to two points because they improve the guide the action is conditioned on, and the ceiling of that mechanism is set by how good a guide the model can write. It also suggests where the offline protocol is most useful, since a mediated reward that pays a guide by the action it induces is computable only when the observation can be replayed under a forced prefix (Appendix~\ref{app:grpo}).

\section{Conclusion}
An offline, deterministic protocol turned a training phenomenon into a measurable object. The mutual reinforcement effect holds in web guide agents: jointly decoding a human oriented guide with the grounded action improves element selection for both Qwen3.5-4B and Qwen3.5-9B, in both decoding orders, and the gain grows with model scale. A forced prefix intervention with a paraphrase control shows the guide is a causal channel rather than commentary: the model executes whichever guide it is given, and a correct one nearly doubles its accuracy. The same channel yields an offline reward no live rollout could compute, though optimizing it brings no gain yet (Appendix~\ref{app:grpo}). On agent data a guide is supervision.

\section*{Limitations}
WebMRE derives from a single benchmark family, so site and layout diversity is bounded by WebArena, and our offline step metrics complement rather than replace live end to end evaluation. The reinforcement learning study uses a deliberately small budget of two rounds. Extending the corpus to further environments and the offline protocol to full trajectories is left to future work.

\bibliography{custom}

\appendix
\setcounter{figure}{0}
\renewcommand{\thefigure}{A\arabic{figure}}
\setcounter{table}{0}
\renewcommand{\thetable}{A\arabic{table}}

\section{Prompt Templates}
\label{app:prompts}
Figures~\ref{fig:prompt-som} to~\ref{fig:prompt-variants} reproduce the locked prompts verbatim. Every training condition, every zero shot baseline, and every judged sample in the paper was rendered from these templates; the only degrees of freedom are the output contract paragraph and the grounding key, exactly as listed in Figure~\ref{fig:prompt-variants}.
\input{tables/appendix_prompts_v2}

\section{Audit and Judge Prompts}
\label{app:audit}
Figures~\ref{fig:prompt-audit-a} to~\ref{fig:prompt-useval} reproduce the judge prompts verbatim: the label audit verdict prompt (Job A), the guide rewrite prompt and its blind pairwise acceptance judge (Job B), and the usefulness rating judge of Section~\ref{sec:channel}. Each judge sees the screenshot and answers in strict JSON; the same prompt goes to all three vendors.
\input{tables/appendix_audit_v2}

\section{Worked Examples}
\label{app:examples}
Figures~\ref{fig:example-1} to~\ref{fig:example-6} show six complete test steps, one per site: the SoM screenshot exactly as the model sees it, the text side of the input, and the verbatim output of the fine tuned 4B JG model against the audited gold action. Four are solved and two are failures; the failures illustrate the coupling of Section~\ref{sec:channel}, where a wrong guide and a wrong action arrive together.
\input{tables/appendix_examples_v2}

\section{Dataset Statistics}
\label{app:stats}
\begin{figure*}[t]
  \centering
  \includegraphics[width=\textwidth]{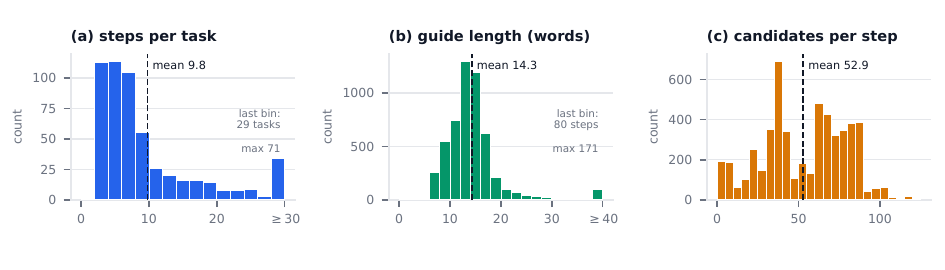}
  \caption{Corpus distributions: steps per task, guide length, and candidate menu size. Dashed lines mark means.}
  \label{fig:dist}
\end{figure*}

Tables~\ref{tab:sitecomp} to~\ref{tab:trainconf} give the full composition of WebMRE and the training configurations; Figure~\ref{fig:dist} shows the distributions behind the corpus level means. Guides are uniformly short (median 13 words), candidate menus are large (median 55 elements), and task length is heavy tailed (median 6 steps, maximum 71), which is why the guide history rather than the raw action history is what later steps consume.
\input{tables/appendix_stats}

\section{Full Results}
\label{app:fullres}
Table~\ref{tab:seeds} gives the per seed values behind every mean reported in the main text, together with the per seed change from the action only reference. Element level scoring excludes the 51 disputed steps throughout; the per site and per flag breakdowns behind Section~\ref{sec:slices} are Tables~\ref{tab:fullsite} and~\ref{tab:fullstale} in the main text.
\input{tables/tableA_seeds}

\section{Mediated-Guide GRPO}
\label{app:grpo}
The causal channel of Section~\ref{sec:channel} suggests a reward: pay a guide by the action it causes. Mediated-Guide GRPO (MG-GRPO) samples $G=8$ completions per step from the guide first model, computes group mean centered advantages without standard deviation scaling \citep{shao2024deepseekmath,liu2025understanding}, drops zero variance groups \citep{NEURIPS2025_a4277440}, and rewards each sample with
\begin{equation}
R_i = F_i\big[(1-\lambda)\,R^{\mathrm{act}}_i + \lambda\,R^{\mathrm{med}}_i\big],
\quad \lambda = 0.3,
\label{eq:reward}
\end{equation}
where $F_i$ and $R^{\mathrm{act}}_i$ are the format gate and verifiable action match of Section~\ref{sec:metrics}, and $R^{\mathrm{med}}_i$ applies the Section~\ref{sec:channel} intervention at training time: the sampled guide is forced as a prefix and earns the correctness of the action it then induces. No reference text, learned reward model, or environment enters the loop, and the mediated term is computable only offline, since it re-queries the policy under a forced prefix on a frozen observation. A control arm sets $\lambda=0$.

We train both arms for two rounds of 1{,}500 prompts on the Qwen3.5-4B guide first checkpoint. Sampling uses temperature $1.0$ and top $p$ $0.95$ with at most 128 new tokens per completion and 64 greedy tokens for the mediated decode; each rank processes 16 prompts per update on four GPUs; advantages are mean centered within each group of eight; checkpoints are saved per round and scored with the frozen protocol of Section~\ref{sec:metrics}. The full trainer is under three hundred lines and touches no environment.

\input{tables/tableA_grpo}

The training reward rises over both rounds, from about .45 to above .80 per chunk in both arms, so the objective is learnable from verifiable step rewards alone. Two things happen on held out data (Table~\ref{tab:grpo}). The mediated arm recovers across rounds and ends above the action only control, .433 against .429, so paying a guide by the action it causes is the better of the two rewards under the same budget. Neither arm, however, improves on the supervised starting point: every checkpoint lands within about one point below .436. We report this plainly as a negative result on a strong starting point. The supervised guide first model is already close to what this backbone does on the task, and a two round budget of 1{,}500 prompts appears to sharpen the sampling distribution rather than change greedy decisions. The contribution here is therefore the reward construction, which the deterministic offline protocol makes computable at all, rather than a gain from optimizing it.

\end{document}

%% file: tables/box_dataexample.tex
\begin{figure}[t]
\centering
\fbox{\begin{minipage}{0.94\columnwidth}\small
\textbf{Two steps of one task} (task 0, shop.\ admin).\\[2pt]
\textit{Intent:} \texttt{What is the top-1 best-selling product in 2022}\\[3pt]
\textit{Step 1 guide:} \texttt{Open the Bestsellers report from the Products section of the Reports menu.}\\
\textit{Step 1 action:} \texttt{click, id 68, (714, 442)}\\[3pt]
\textit{Step 3 guide:} \texttt{Enter 01/01/22 in the From date field to set the report start date.}\\
\textit{Step 3 action:} \texttt{type, id 70, (498, 511), ``01/01/22''}
\end{minipage}}
\caption{Two audited WebMRE steps. Each record carries the task intent, a guide sentence, and one action grounded both as a marked element id and as pixel coordinates; full records in Appendix~\ref{app:examples}.}
\label{fig:dataexample}
\end{figure}

%% file: tables/table2_main.tex
\begin{table}[t]
\centering
\small
\begin{tabular}{llcc}
\toprule
Model & Contract & SAA & BLEU-1 \\
\midrule
\multicolumn{4}{l}{\textit{Qwen3.5-4B (fine tuned)}} \\
\addlinespace[1pt]
& A & \textbf{.423} & -- \\
& J & .432 & .551 \\
& JG & .425 & .545 \\
\addlinespace[3pt]
\multicolumn{4}{l}{\textit{Qwen3.5-9B (fine tuned)}} \\
\addlinespace[1pt]
& A & .404 & -- \\
& G & -- & \textbf{.563} \\
& J & .421 & .551 \\
& JG & \textbf{.426} & .558 \\
\addlinespace[3pt]
\multicolumn{4}{l}{\textit{Frontier models (zero shot, contract J)}} \\
\addlinespace[1pt]
GPT-5.5 & J & .259 & .359 \\
Claude Opus 4.8 & J & .261 & .363 \\
Gemini 3.5 Flash & J & .204 & .238 \\
\bottomrule
\end{tabular}
\caption{Main results on the SoM view (element selection). SAA is the offline step accuracy of Section~\ref{sec:metrics}; BLEU-1 scores the generated guide against the gold guide. Fine tuned rows are the mean over three seeds. Both joint contracts (J, JG) raise SAA over the action only reference (A) at both model sizes; A and G rows have no guide or no action output by contract.}
\label{tab:main}
\end{table}

%% file: tables/table_multiseed.tex
\begin{table}[t]
\centering
\small
\begin{tabular}{lcccc}
\toprule
Model & A & J & JG & $\Delta$(J,\,JG) vs A \\
\midrule
Qwen3.5-4B & .423 & .432 & .425 & $+0.9,\ +0.2$ \\
Qwen3.5-9B & .404 & .421 & .426 & $+1.7,\ +2.2$ \\
\bottomrule
\end{tabular}
\caption{The mutual reinforcement effect on SoM element selection, as the mean SAA over three training seeds. Joint guide and action training (J action first, JG guide first) improves accuracy over the action only reference (A) for both models, and the improvement grows with model scale: from $+0.9$ / $+0.2$ points at 4B to $+1.7$ / $+2.2$ points at 9B.}
\label{tab:multiseed}
\end{table}

%% file: tables/appendix_fullresults.tex
\begin{table}[t]
\centering
\footnotesize
\setlength{\tabcolsep}{3.5pt}
\begin{tabular}{llcccccc}
\toprule
Model & C. & gitlab & map & reddit & shop & s.adm & wiki \\
\midrule
4B & A & .502 & .597 & .528 & .494 & .266 & .667 \\
4B & J & .512 & .609 & .506 & .500 & .281 & .727 \\
4B & JG & .542 & .600 & .525 & .403 & .278 & .697 \\
\addlinespace[3pt]
9B & A & .482 & .562 & .514 & .457 & .262 & .485 \\
9B & J & .523 & .592 & .561 & .418 & .271 & .485 \\
9B & JG & .514 & .585 & .522 & .449 & .292 & .515 \\
\bottomrule
\end{tabular}
\caption{Step accuracy by site, pooled over the three seeds. A joint contract is the best row on five of the six sites for each model, the exception being reddit at 4B and shopping at 9B, so the mutual reinforcement effect is not carried by a single site.}
\label{tab:fullsite}
\end{table}

\begin{table}[t]
\centering
\small
\begin{tabular}{llcc}
\toprule
Model & Contract & flagged & unflagged \\
\midrule
4B & A & .399 & .472 \\
4B & J & .400 & .498 \\
4B & JG & .409 & .459 \\
\addlinespace[3pt]
9B & A & .372 & .472 \\
9B & J & .392 & .480 \\
9B & JG & .402 & .476 \\
\bottomrule
\end{tabular}
\caption{Step accuracy split by the per step obs\_stale flag of Section~\ref{sec:provenance}, pooled over the three seeds. Flagged steps are harder for every contract. Both joint contracts gain on the flagged slice at both sizes, and on the unflagged slice they hold at or above the action only reference except for 4B guide first, so the recapture disclosure does not drive the effect.}
\label{tab:fullstale}
\end{table}

%% file: tables/table_channel.tex
\begin{table}[t]
\centering
\small
\setlength{\tabcolsep}{4pt}
\begin{tabular}{llc}
\toprule
Forced prefix & What it tests & SAA \\
\midrule
none, free run & baseline & .422 \\
own guide & intervention neutral & .422 \\
\addlinespace[3pt]
gold guide & a correct guide & \textbf{.684} \\
paraphrased gold & meaning, not overlap & .554 \\
another step's guide & a wrong guide & .055 \\
\bottomrule
\end{tabular}
\caption{The guide is a causal channel. On all 1{,}220 SoM test steps we force the decoder to open with a chosen guide and let it complete only the action, on the Qwen3.5-9B guide first model. Forcing the model's own guide reproduces free decoding, so the intervention itself is neutral; a correct guide then lifts action accuracy by 26 points and a guide belonging to a different step destroys it. The paraphrase, which names the target by function rather than by its on screen label, recovers half of that gain.}
\label{tab:channel}
\end{table}

%% file: tables/appendix_prompts_v2.tex

\begin{figure*}[p]
\begin{promptbox}{System Prompt (SoM view, contract J), verbatim}
\begin{multicols}{2}
{\scriptsize\ttfamily\raggedright
You are a vision web agent. You operate a real website one step at a time to accomplish a user's task, and at every step you also produce one short piece of in-app guidance for a human who is doing the same step.\par
\smallskip
At each step you are given:\par
1. The user's task (the goal to accomplish).\par
2. A screenshot of the current page. Interactive elements are marked with numbered boxes (Set-of-Marks); each number is the id of one candidate element.\par
3. A numbered list of candidate elements, one per line, formatted as: [id] TYPE visible text  (for example: [17] BUTTON Add to Cart). Each id matches a numbered box in the screenshot.\par
4. The history of guidance already produced this episode (what has been done so far).\par
\smallskip
Decide the single best NEXT action, then write its guide\_text.\par
\smallskip
\textbf{OUTPUT CONTRACT}\par
Return exactly one valid JSON object and nothing else: no markdown, no code fences, no text before or after. The object must have exactly these keys, in this order: ''action\_type'', ''selected\_candidate\_id'', ''content'', ''guide\_text''.\par
\textbullet\ ''action\_type'': exactly one of: ''click'', ''type'', ''select'', ''scroll'', ''press\_enter'', ''go\_back'', ''finish''.\par
\textbullet\ ''selected\_candidate\_id'': the id (as a string) of the target element from the candidate list, or null.\par
\textbullet\ ''content'': the action's payload, or null (see the per-action rules below).\par
\textbullet\ ''guide\_text'': one short instruction (one sentence) telling a HUMAN what to do this step.\par
\smallskip
\textbf{ACTION SPACE — use only these seven action types, and follow each parameter rule exactly:}\par
\textbullet\ ''click'': selected\_candidate\_id = the target id; content = null.\par
\textbullet\ ''type'': selected\_candidate\_id = the input field id; content = the exact text to type.\par
\textbullet\ ''select'': selected\_candidate\_id = the dropdown id; content = the exact visible option label to choose.\par
\textbullet\ ''scroll'': selected\_candidate\_id = null; content = ''up'' or ''down''.\par
\textbullet\ ''press\_enter'': selected\_candidate\_id = null; content = null.\par
\textbullet\ ''go\_back'': selected\_candidate\_id = null; content = null.\par
\textbullet\ ''finish'': selected\_candidate\_id = null; content = the final answer if the task asks for one, otherwise null.\par
\textbullet\ For ''click'', ''type'', and ''select'', selected\_candidate\_id MUST be one of the ids present in the candidate list. Never invent an id.\par
\smallskip
\textbf{guide\_text RULES:}\par
\textbullet\ guide\_text speaks to the human user in plain language (''Click...'', ''Type ... into the search box'', ''Open the Reports menu''), and refers to the target by its visible on-screen label or location.\par
\textbullet\ guide\_text must NOT contain candidate ids, numbers from the marks, coordinates, bounding boxes, DOM, CSS, or selector details.\par
\textbullet\ guide\_text is one short direct instruction, not reasoning or a prediction.\par
\smallskip
Return exactly one JSON object with keys action\_type, selected\_candidate\_id, content, guide\_text and nothing else.\par
}
\end{multicols}
\end{promptbox}
\caption{The locked system prompt of the SoM view under contract J, verbatim.}
\label{fig:prompt-som}
\end{figure*}

\begin{figure*}[p]
\begin{promptbox}{System Prompt (coordinate view, contract J), verbatim}
\begin{multicols}{2}
{\scriptsize\ttfamily\raggedright
You are a vision web agent. You operate a real website one step at a time to accomplish a user's task, and at every step you also produce one short piece of in-app guidance for a human who is doing the same step.\par
\smallskip
At each step you are given:\par
1. The user's task (the goal to accomplish).\par
2. A screenshot of the current page. Interactive elements are marked with numbered boxes (Set-of-Marks); each number is the id of one candidate element.\par
3. A numbered list of candidate elements, one per line, formatted as: [id] TYPE visible text  (for example: [17] BUTTON Add to Cart). Each id matches a numbered box in the screenshot.\par
4. The history of guidance already produced this episode (what has been done so far).\par
\smallskip
Decide the single best NEXT action, then write its guide\_text.\par
\smallskip
\textbf{OUTPUT CONTRACT}\par
Return exactly one valid JSON object and nothing else: no markdown, no code fences, no text before or after. The object must have exactly these keys, in this order: ''action\_type'', ''coords'', ''content'', ''guide\_text''.\par
\textbullet\ ''action\_type'': exactly one of: ''click'', ''type'', ''select'', ''scroll'', ''press\_enter'', ''go\_back'', ''finish''.\par
\textbullet\ ''coords'': [x, y] integer pixel position of the target element, or null.\par
\textbullet\ ''content'': the action's payload, or null (see the per-action rules below).\par
\textbullet\ ''guide\_text'': one short instruction (one sentence) telling a HUMAN what to do this step.\par
\smallskip
\textbf{ACTION SPACE — use only these seven action types, and follow each parameter rule exactly:}\par
\textbullet\ ''click'': coords = [x, y], the pixel position of the target element's center; content = null.\par
\textbullet\ ''type'': coords = [x, y] of the input field; content = the exact text to type.\par
\textbullet\ ''select'': coords = [x, y] of the dropdown; content = the exact visible option label to choose.\par
\textbullet\ ''scroll'': coords = null; content = ''up'' or ''down''.\par
\textbullet\ ''press\_enter'': coords = null; content = null.\par
\textbullet\ ''go\_back'': coords = null; content = null.\par
\textbullet\ ''finish'': coords = null; content = the final answer if the task asks for one, otherwise null.\par
\textbullet\ coords are integer pixel positions on the screenshot (origin top-left); aim at the center of the target element.\par
\smallskip
\textbf{guide\_text RULES:}\par
\textbullet\ guide\_text speaks to the human user in plain language (''Click...'', ''Type ... into the search box'', ''Open the Reports menu''), and refers to the target by its visible on-screen label or location.\par
\textbullet\ guide\_text must NOT contain candidate ids, numbers from the marks, coordinates, bounding boxes, DOM, CSS, or selector details.\par
\textbullet\ guide\_text is one short direct instruction, not reasoning or a prediction.\par
\smallskip
Return exactly one JSON object with keys action\_type, coords, content, guide\_text and nothing else.\par
}
\end{multicols}
\end{promptbox}
\caption{The locked system prompt of the coordinate view under contract J, verbatim.}
\label{fig:prompt-coord}
\end{figure*}

\begin{figure*}[p]
\begin{promptbox}{Contract and view variants, and the user message template, verbatim}
\begin{multicols}{2}
{\scriptsize\ttfamily\raggedright
\textbf{Contract variants.}\par
\smallskip
The other contracts change only the OUTPUT CONTRACT paragraph and the final instruction line. A drops guide\_text and its rules; G drops the action keys and the action space; JG lists guide\_text before the action keys. The coordinate view replaces selected\_candidate\_id with coords = [x, y] and swaps the per action parameter rules accordingly. Final line per contract:\par
\smallskip
\textbf{som:A}\ \ Decide the single best NEXT action. Return ONLY one JSON object with exactly the keys: action\_type, selected\_candidate\_id, content.\par
\smallskip
\textbf{som:G}\ \ Write the guide\_text for the NEXT step a human should take. Return ONLY one JSON object with exactly the keys: guide\_text.\par
\smallskip
\textbf{som:J}\ \ Decide the single best NEXT action, then write its guide\_text. Return ONLY one JSON object with exactly the keys: action\_type, selected\_candidate\_id, content, guide\_text.\par
\smallskip
\textbf{som:JG}\ \ First write the guide\_text for the next step, then give the corresponding action. Return ONLY one JSON object with exactly the keys: guide\_text, action\_type, selected\_candidate\_id, content.\par
\smallskip
\textbf{coord:A}\ \ Decide the single best NEXT action. Return ONLY one JSON object with exactly the keys: action\_type, coords, content.\par
\smallskip
\textbf{coord:G}\ \ Write the guide\_text for the NEXT step a human should take. Return ONLY one JSON object with exactly the keys: guide\_text.\par
\smallskip
\textbf{coord:J}\ \ Decide the single best NEXT action, then write its guide\_text. Return ONLY one JSON object with exactly the keys: action\_type, coords, content, guide\_text.\par
\smallskip
\textbf{coord:JG}\ \ First write the guide\_text for the next step, then give the corresponding action. Return ONLY one JSON object with exactly the keys: guide\_text, action\_type, coords, content.\par
\smallskip
\textbf{User template.}\par
\smallskip
Task:\par
\{intent\}\par
\smallskip
Action history (what has been done so far; oldest first, empty if this is the first step):\par
\{history\}\par
\smallskip
Candidate elements on the current page (each line is ''[id] TYPE visible text''):\par
\{candidates\}\par
\smallskip
The current page screenshot, with numbered boxes on the candidate elements matching the ids above, is attached as an image.\par
\smallskip
\{final line of the contract\}\par
}
\end{multicols}
\end{promptbox}
\caption{What changes between contracts and views, the final instruction line of every contract, and the user message template, all verbatim.}
\label{fig:prompt-variants}
\end{figure*}

%% file: tables/appendix_audit_v2.tex

\begin{figure*}[tp]
\begin{promptbox}{Label audit judge (Job A), verbatim}
\begin{multicols}{2}
{\scriptsize\ttfamily\raggedright
You are auditing a web-agent dataset. Each sample pairs a Set-of-Mark (SoM)\par
screenshot with a recorded ''gold'' element id for one UI action. The recorded id\par
was produced by an automatic coordinate-to-mark mapping and may be wrong.\par
\smallskip
Decide whether the recorded element id is the element a human would interact\par
with to follow the given step instruction (guide) toward the task intent.\par
\smallskip
\textbf{Rules:}\par
\textbullet\ Judge ONLY from the screenshot, the candidate menu, and the click coordinates.\par
\textbullet\ ''confirm'': the recorded id clearly matches the guide's target element.\par
\textbullet\ ''remap'': a DIFFERENT candidate id clearly matches the guide and is consistent\par
with the click coordinates (same region). Give that id.\par
\textbullet\ ''unmappable'': no candidate in the menu corresponds to the guide's target\par
(e.g. the element was not tagged). Do not force a nearest guess.\par
\textbullet\ When uncertain between confirm and remap, prefer ''confirm'' (conservative).\par
\smallskip
Output strict JSON:\par
\{''verdict'': ''confirm'' | ''remap'' | ''unmappable'',\par
''candidate\_id'': ''<id or null>'',\par
''reason'': ''<one short sentence>''\}\par
}
\end{multicols}
\end{promptbox}
\caption{The system prompt of the label audit judge in Job A, verbatim.}
\label{fig:prompt-audit-a}
\vspace{1.2em}
\begin{promptbox}{Guide rewrite (Job B), verbatim}
\begin{multicols}{2}
{\scriptsize\ttfamily\raggedright
You are writing the in-app guide text for ONE step of a web task, as shown to\par
an end user by a digital-adoption tool (like an overlay tooltip).\par
\smallskip
\textbf{Requirements:}\par
\textbullet\ Imperative voice, ONE action per step, <= 25 words.\par
\textbullet\ Name the target element by its VISIBLE text or an unambiguous visual\par
description; never invent labels that are not on the screen.\par
\textbullet\ Never mention element ids, marks, numbers in brackets, or pixel coordinates.\par
\textbullet\ For ''type'' steps, state what to enter and where.\par
\textbullet\ Be consistent with the task intent and the prior steps' guides.\par
\textbullet\ The action shown below is the ground truth of what this step does; describe\par
exactly that action, nothing else.\par
\smallskip
Output strict JSON: \{''guide\_text'': ''<the rewritten guide>''\}\par
}
\end{multicols}
\end{promptbox}
\caption{The system prompt of the guide rewrite step in Job B, verbatim.}
\label{fig:prompt-audit-b}
\end{figure*}

\begin{figure*}[tp]
\begin{promptbox}{Blind pairwise usefulness judge (Job B acceptance), verbatim}
\begin{multicols}{2}
{\scriptsize\ttfamily\raggedright
You rate in-app step guides for a web task. A guide is shown as a tooltip to an\par
end user on the given screenshot. Rate each of the two guides below independently:\par
\textbullet\ ''useful'': correctly and unambiguously tells the user what to do for THIS step\par
(the ground-truth action is given), names a real on-screen target.\par
\textbullet\ ''ambiguous'': partially right but vague, or the target is unclear.\par
\textbullet\ ''useless'': wrong action, wrong/invented target, or misleading.\par
\smallskip
Output strict JSON: \{''a'': ''useful''|''ambiguous''|''useless'', ''b'': ''useful''|''ambiguous''|''useless''\}\par
}
\end{multicols}
\end{promptbox}
\caption{The system prompt of the blind pairwise usefulness judge that gates Job B rewrites, verbatim.}
\label{fig:prompt-audit-judge}
\vspace{1.2em}
\begin{promptbox}{Usefulness rating judge, verbatim}
\begin{multicols}{2}
{\scriptsize\ttfamily\raggedright
You rate ONE in-app step guide for a web task. The guide is shown as a\par
tooltip to an end user on the given screenshot. You are told the ground-truth\par
action of this step. Rate the guide:\par
\textbullet\ ''useful'': correctly and unambiguously tells the user what to do for THIS step,\par
names a real on-screen target consistent with the ground-truth action.\par
\textbullet\ ''ambiguous'': partially right but vague, or the target is unclear.\par
\textbullet\ ''useless'': wrong action, wrong/invented target, or misleading.\par
Output strict JSON: \{''rating'': ''useful'' | ''ambiguous'' | ''useless''\}\par
}
\end{multicols}
\end{promptbox}
\caption{The system prompt of the usefulness rating judge for model generated guides, verbatim.}
\label{fig:prompt-useval}
\end{figure*}

%% file: tables/appendix_examples_v2.tex

\begin{figure*}[p]
\centering
\begin{promptbox}{Worked example 1 of 6: gitlab, task 315, step 3, gold action click, model CORRECT}
\centering
\includegraphics[width=0.88\textwidth,height=0.52\textheight,keepaspectratio]{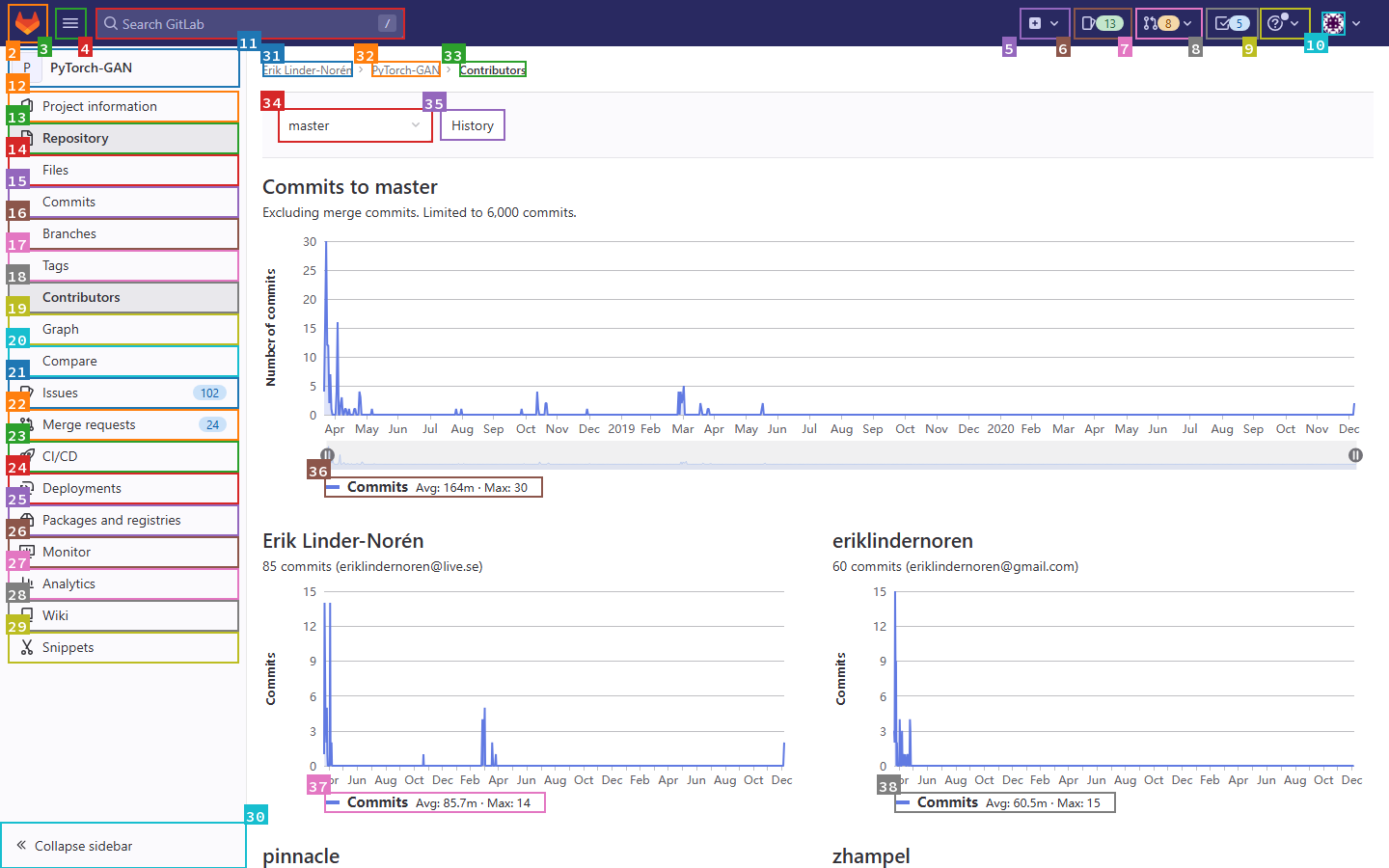}
\tcblower
\begin{minipage}[t]{0.48\textwidth}{\scriptsize\ttfamily\raggedright
\textbf{Model input (text part)}\par\smallskip
Task intent: List the email address of the top 3 contributors to Pytorch GAN repo, ranked by the number of commits?\par
History (last 2 of 3 prior guides):\par
- Open the PyTorch-GAN project from the search results.\par
- Click the Repository section in the left sidebar to access repository tools like contributors.\par
Candidates: 37 elements; 3 shown:\par
[13] A Repository\par
[18] A Contributors\par
[33] A Contributors\par
...\par
}\end{minipage}\hfill
\begin{minipage}[t]{0.48\textwidth}{\scriptsize\ttfamily\raggedright
\textbf{Model output (4B, contract JG, verbatim)}\par\smallskip
\{''guide\_text'': ''Click Contributors in the Repository menu to view the contributor commit counts.'', ''action\_type'': ''click'', ''selected\_candidate\_id'': ''18'', ''content'': null\}\par\smallskip
\textbf{Gold action:} type=click, id=18, content=null\par
\textbf{Result:} correct; the model clicks the same Contributors entry (candidate 18) as the gold action.\par
}\end{minipage}
\end{promptbox}
\caption{Worked example 1. The model opens the Contributors page of the PyTorch GAN repository by clicking candidate 18, matching the gold action.}
\label{fig:example-1}
\end{figure*}

\begin{figure*}[p]
\centering
\begin{promptbox}{Worked example 2 of 6: map, task 20, step 4, gold action select, model CORRECT}
\centering
\includegraphics[width=0.88\textwidth,height=0.52\textheight,keepaspectratio]{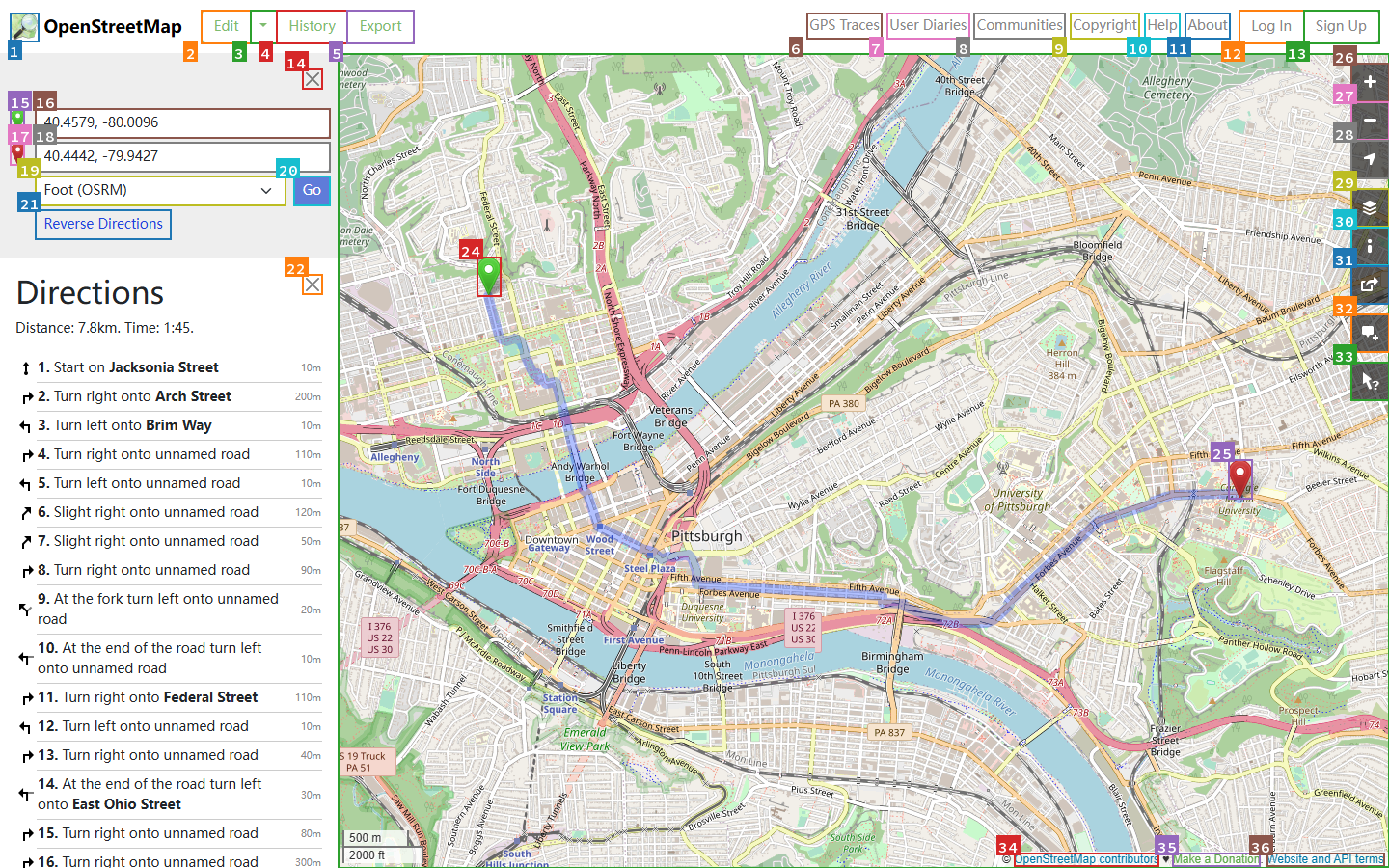}
\tcblower
\begin{minipage}[t]{0.48\textwidth}{\scriptsize\ttfamily\raggedright
\textbf{Model input (text part)}\par\smallskip
Task intent: Compare the difference in time for walking and driving route from Randyland to Carnegie Mellon University\par
History (last 2 of 4 prior guides):\par
- Enter “Carnegie Mellon University” in the To field.\par
- Click Go to calculate the driving route.\par
Candidates: 36 elements; 3 shown:\par
[1] IMG OpenStreetMap logo\par
[2] A Edit\par
[19] SELECT Bicycle (OSRM)Car (OSRM)Foot (OSRM)\par
...\par
}\end{minipage}\hfill
\begin{minipage}[t]{0.48\textwidth}{\scriptsize\ttfamily\raggedright
\textbf{Model output (4B, contract JG, verbatim)}\par\smallskip
\{''guide\_text'': ''Open the travel mode dropdown and choose “Car (OSRM)” to switch to the driving route.'', ''action\_type'': ''select'', ''selected\_candidate\_id'': ''19'', ''content'': ''Car (OSRM)''\}\par\smallskip
\textbf{Gold action:} type=select, id=19, content='Car (OSRM)'\par
\textbf{Result:} correct; action type select, candidate 19, and the option Car (OSRM) all match the gold action.\par
}\end{minipage}
\end{promptbox}
\caption{Worked example 2. The model selects the Car (OSRM) option in the travel mode dropdown, matching the gold action of the route comparison task.}
\label{fig:example-2}
\end{figure*}

\begin{figure*}[p]
\centering
\begin{promptbox}{Worked example 3 of 6: reddit, task 69, step 3, gold action finish, model CORRECT}
\centering
\includegraphics[width=0.88\textwidth,height=0.52\textheight,keepaspectratio]{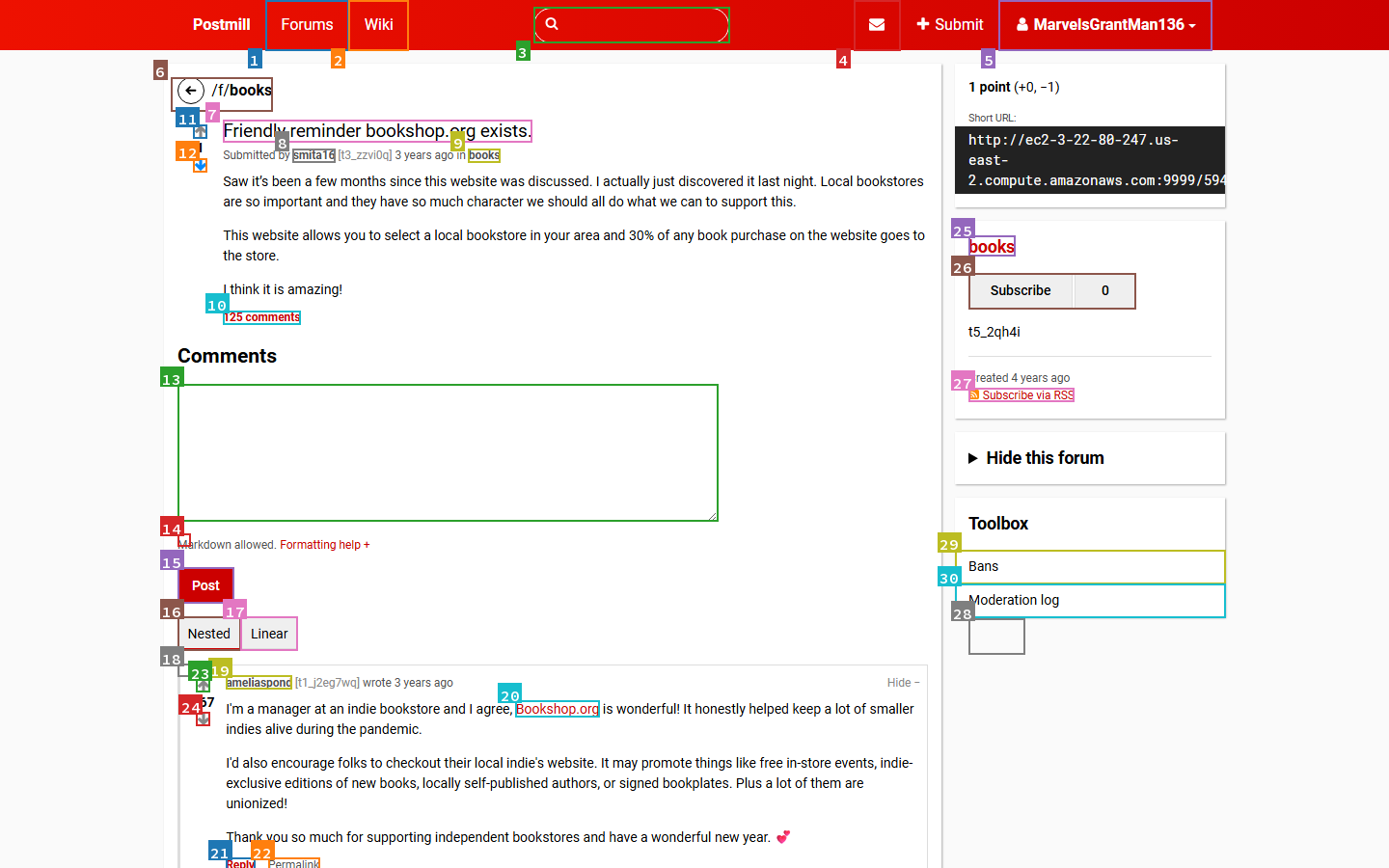}
\tcblower
\begin{minipage}[t]{0.48\textwidth}{\scriptsize\ttfamily\raggedright
\textbf{Model input (text part)}\par\smallskip
Task intent: Among the top 10 post in ''books'' forum, is there any post talks about supporting local book stores? If so, tell me the organizations involved\par
History (last 2 of 2 prior guides):\par
- Open the comments for this post so you can verify details about supporting local bookstores and i...\par
- Scroll down through the comments to look for mentions of organizations connected to supporting lo...\par
Candidates: 30 elements; 3 shown:\par
[7] A Friendly reminder bookshop.org exists.\par
[9] A books\par
[20] A Bookshop.org\par
...\par
}\end{minipage}\hfill
\begin{minipage}[t]{0.48\textwidth}{\scriptsize\ttfamily\raggedright
\textbf{Model output (4B, contract JG, verbatim)}\par\smallskip
\{''guide\_text'': ''You're done — the answer is: Yes, the post 'Friendly reminder bookshop.org exists.' mentions supporting local bookstores. The organization involved is Bookshop.org.'', ''action\_type'': ''finish'', ''selected\_candidate\_id'': null, ''content'': ''Yes, the post 'Friendly reminder bookshop.org exists.' mentions supporting local bookstores. The organization involved is Bookshop.org.''\}\par\smallskip
\textbf{Gold action:} type=finish, id=null, content='Yes, the post titled ''Friendly reminder bookshop.org exists.'' discusses supporting local bookstor...'\par
\textbf{Result:} correct; the model issues finish and its answer names the same post and organization, Bookshop.org, as the gold answer.\par
}\end{minipage}
\end{promptbox}
\caption{Worked example 3. The model finishes the forum question with an answer naming the Bookshop.org post, matching the gold answer.}
\label{fig:example-3}
\end{figure*}

\begin{figure*}[p]
\centering
\begin{promptbox}{Worked example 4 of 6: shopping, task 672, step 3, gold action scroll, model INCORRECT}
\centering
\includegraphics[width=0.88\textwidth,height=0.52\textheight,keepaspectratio]{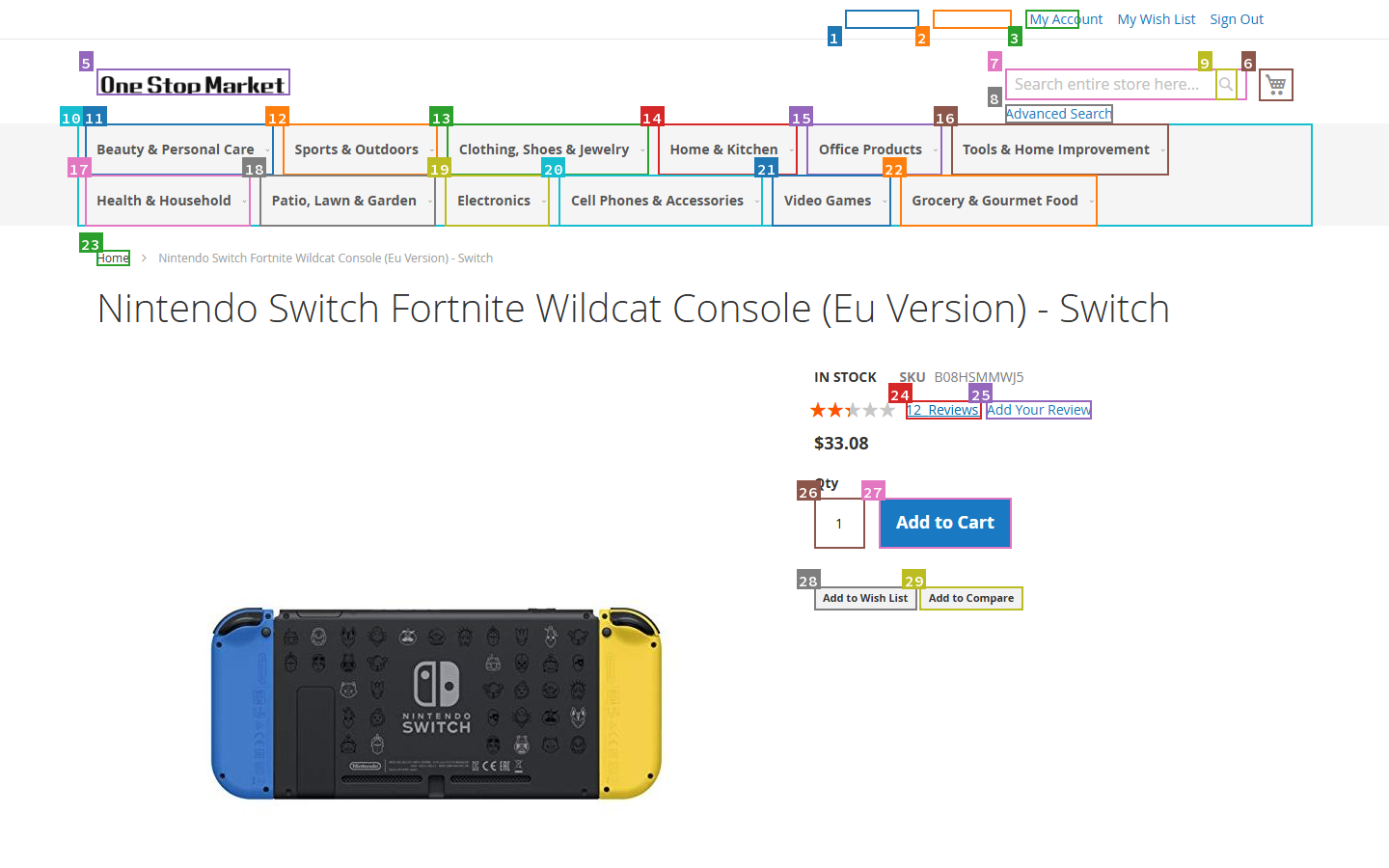}
\tcblower
\begin{minipage}[t]{0.48\textwidth}{\scriptsize\ttfamily\raggedright
\textbf{Model input (text part)}\par\smallskip
Task intent: Gather the titles of Nintendo Switch Fortnite Wildcat Console EU reviews with 3 stars and less rating from OneStopShop, and post them in the games subreddit under the title ''real user feedback on Nintendo Switch Fortnite Wildcat Console EU''\par
History (last 2 of 3 prior guides):\par
- Open the product’s reviews by selecting the reviews link.\par
- Click the “12 Reviews” link below the product title to open the customer reviews.\par
Candidates: 28 elements; 3 shown:\par
[1] A My Account\par
[2] A My Wish List\par
[24] A 12 Reviews\par
...\par
}\end{minipage}\hfill
\begin{minipage}[t]{0.48\textwidth}{\scriptsize\ttfamily\raggedright
\textbf{Model output (4B, contract JG, verbatim)}\par\smallskip
\{''guide\_text'': ''Click the “12 Reviews” link next to the star rating to open the customer reviews.'', ''action\_type'': ''click'', ''selected\_candidate\_id'': ''24'', ''content'': null\}\par\smallskip
\textbf{Gold action:} type=scroll, id=null, content='down'\par
\textbf{Result:} incorrect; gold scrolls down toward the review section, while the model clicks the 12 Reviews link (candidate 24) under an otherwise plausible guide.\par
}\end{minipage}
\end{promptbox}
\caption{Worked example 4. The gold action scrolls down toward the review section, while the model clicks the 12 Reviews link, so the step is scored incorrect despite a plausible guide.}
\label{fig:example-4}
\end{figure*}

\begin{figure*}[p]
\centering
\begin{promptbox}{Worked example 5 of 6: shopping\_admin, task 0, step 0, gold action click, model INCORRECT}
\centering
\includegraphics[width=0.88\textwidth,height=0.52\textheight,keepaspectratio]{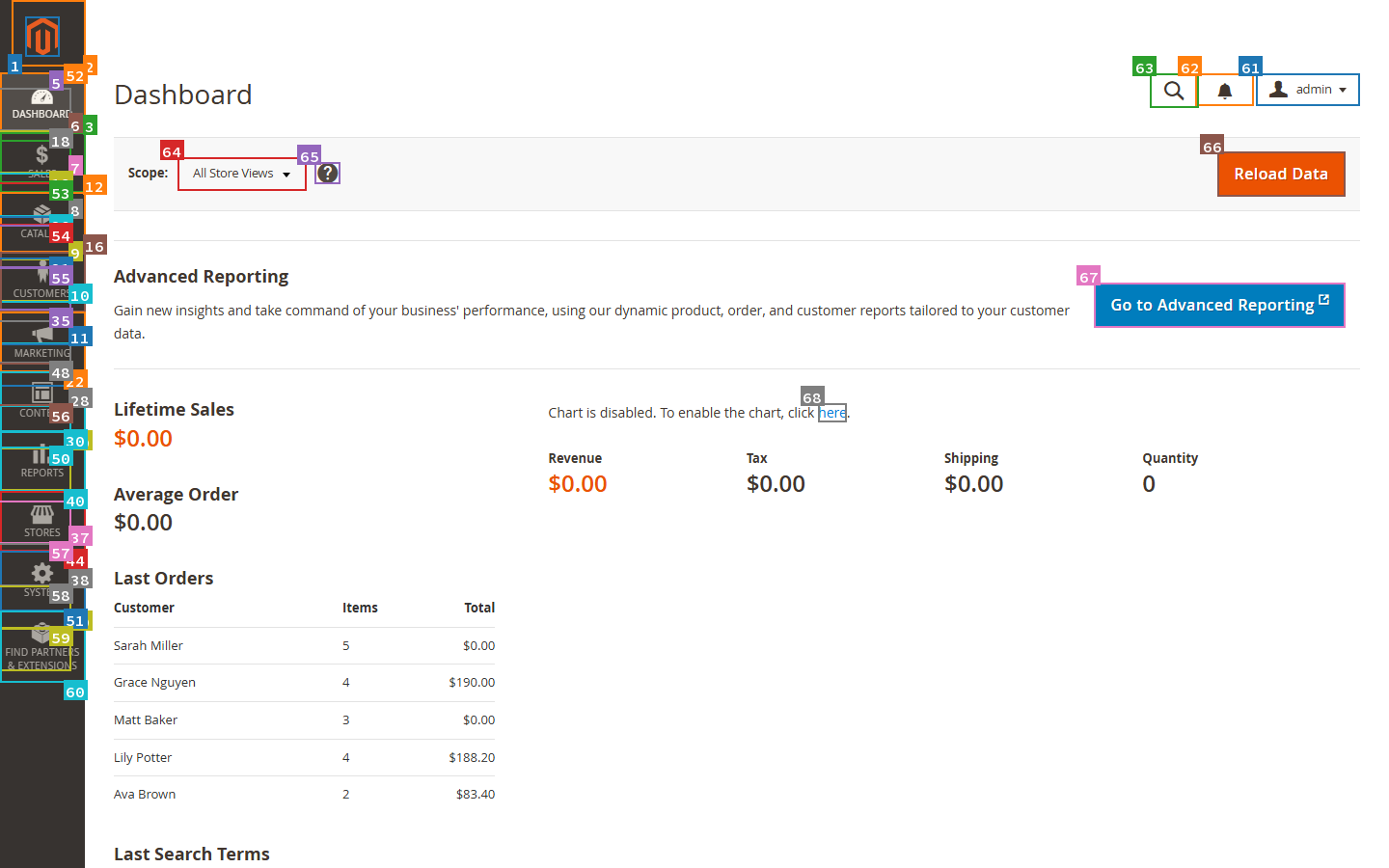}
\tcblower
\begin{minipage}[t]{0.48\textwidth}{\scriptsize\ttfamily\raggedright
\textbf{Model input (text part)}\par\smallskip
Task intent: What is the top-1 best-selling product in 2022\par
History: (first step)\par
Candidates: 61 elements; 3 shown:\par
[1] IMG Magento Admin Panel\par
[29] A Pending Reviews\par
[40] A Reports\par
...\par
}\end{minipage}\hfill
\begin{minipage}[t]{0.48\textwidth}{\scriptsize\ttfamily\raggedright
\textbf{Model output (4B, contract JG, verbatim)}\par\smallskip
\{''guide\_text'': ''Open the Reports section from the left sidebar.'', ''action\_type'': ''click'', ''selected\_candidate\_id'': ''29'', ''content'': null\}\par\smallskip
\textbf{Gold action:} type=click, id=40, content=null\par
\textbf{Result:} incorrect; the guide names the Reports menu, but the selected candidate 29 is Pending Reviews rather than the Reports entry (candidate 40).\par
}\end{minipage}
\end{promptbox}
\caption{Worked example 5. The model writes a guide that names the Reports menu but clicks the Pending Reviews entry, candidate 29 instead of candidate 40, so the step is scored incorrect.}
\label{fig:example-5}
\end{figure*}

\begin{figure*}[p]
\centering
\begin{promptbox}{Worked example 6 of 6: wikipedia, task 268, step 0, gold action type, model CORRECT}
\centering
\includegraphics[width=0.88\textwidth,height=0.52\textheight,keepaspectratio]{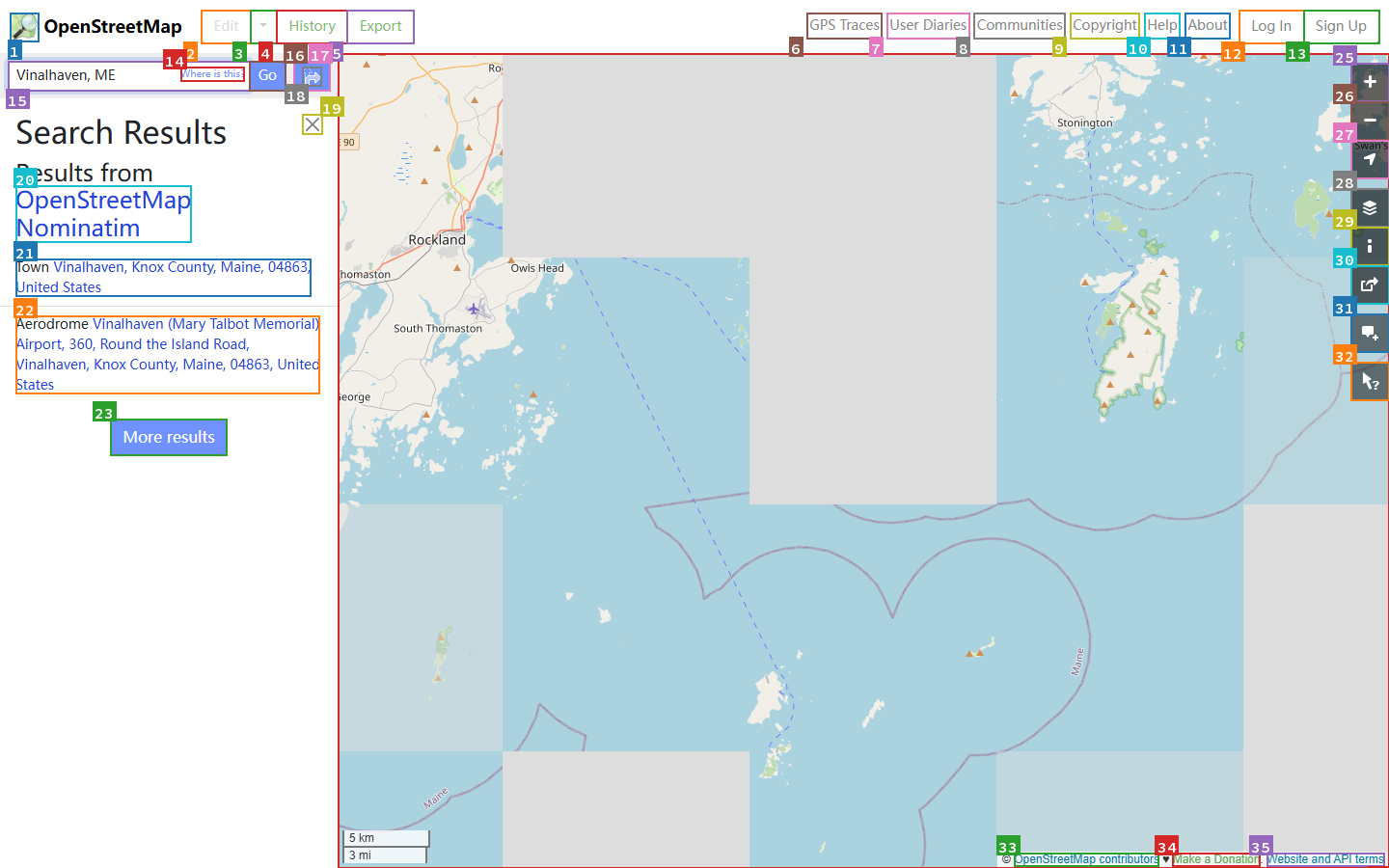}
\tcblower
\begin{minipage}[t]{0.48\textwidth}{\scriptsize\ttfamily\raggedright
\textbf{Model input (text part)}\par\smallskip
Task intent: What's the closest national park to Vinalhaven, ME? How long does it take to bike there?\par
History: (first step)\par
Candidates: 35 elements; 3 shown:\par
[1] IMG OpenStreetMap logo\par
[2] A Edit\par
[15] INPUT\par
...\par
}\end{minipage}\hfill
\begin{minipage}[t]{0.48\textwidth}{\scriptsize\ttfamily\raggedright
\textbf{Model output (4B, contract JG, verbatim)}\par\smallskip
\{''guide\_text'': ''Type “Vinalhaven, ME” into the search box to find the location on the map.'', ''action\_type'': ''type'', ''selected\_candidate\_id'': ''15'', ''content'': ''Vinalhaven, ME''\}\par\smallskip
\textbf{Gold action:} type=type, id=15, content='Vinalhaven, ME'\par
\textbf{Result:} correct; the model types Vinalhaven, ME into the search box (candidate 15), matching the gold action.\par
}\end{minipage}
\end{promptbox}
\caption{Worked example 6. On the first step of a wikipedia labeled task the model types the town name into the map search box, matching the gold action.}
\label{fig:example-6}
\end{figure*}

%% file: tables/appendix_stats.tex
\begin{table}[t]
\centering
\small
\setlength{\tabcolsep}{4pt}
\begin{tabular}{lrrr}
\toprule
Site & Tasks & SoM steps & Coord.\ steps \\
\midrule
gitlab & 86 / 34 & 662 / 303 & 855 / 429 \\
map & 53 / 23 & 464 / 135 & 471 / 135 \\
reddit & 69 / 29 & 363 / 124 & 423 / 160 \\
shopping & 77 / 32 & 423 / 163 & 527 / 225 \\
shop.\ admin & 88 / 38 & 1,115 / 484 & 1,395 / 603 \\
wikipedia & 9 / 3 & 59 / 11 & 59 / 11 \\
\midrule
Total & 382 / 159 & 3,086 / 1,220 & 3,730 / 1,563 \\
\bottomrule
\end{tabular}
\caption{Per site composition. Each cell gives train / test counts under the task level split, which is shared by the two views; every task contributes steps to both views.}
\label{tab:sitecomp}
\end{table}

\begin{table}[t]
\centering
\small
\begin{tabular}{lrrrr}
\toprule
 & \multicolumn{2}{c}{SoM} & \multicolumn{2}{c}{Coordinate} \\
\cmidrule(lr){2-3} \cmidrule(lr){4-5}
Action type & train & test & train & test \\
\midrule
click & 1,622 & 650 (53.3) & 2,140 & 936 (59.9) \\
type & 654 & 282 (23.1) & 772 & 333 (21.3) \\
finish & 382 & 159 (13.0) & 382 & 159 (10.2) \\
scroll & 312 & 93 (7.6) & 312 & 93 (6.0) \\
select & 91 & 28 (2.3) & 99 & 34 (2.2) \\
go\_back & 25 & 8 (0.7) & 25 & 8 (0.5) \\
\bottomrule
\end{tabular}
\caption{Gold action type distribution per view and split. Test columns additionally give the share of the split in percent (parentheses).}
\label{tab:atypes}
\end{table}

\begin{table}[t]
\centering
\small
\begin{tabular}{lrrrr}
\toprule
Metric & Mean & Median & P90 & Max \\
\midrule
Guide length (words) & 14.3 & 13 & 18 & 171 \\
Intent length (words) & 13.6 & 13 & -- & 42 \\
Candidates per step & 52.9 & 55 & -- & 127 \\
Steps per task & 9.8 & 6 & -- & 71 \\
\bottomrule
\end{tabular}
\caption{Text and structure statistics on the unified corpus (5,293 steps, 541 tasks). Guide and candidate statistics are per step; intent and step count statistics are per task. P90 is the 90th percentile.}
\label{tab:textstats}
\end{table}

\begin{table}[t]
\centering
\footnotesize
\setlength{\tabcolsep}{2.5pt}
\begin{tabular}{lccp{2.8cm}}
\toprule
Family & Ep.\,/\,rd. & LR & Other \\
\midrule
Qwen3.5-4B SFT & 3 ep. & $10^{-5}$ & \raggedright batch 8, cosine, warmup 0.03, seeds 20260731 to 20260733 \tabularnewline
\addlinespace[2pt]
Qwen3.5-9B SFT & 3 ep. & $10^{-5}$ & \raggedright as 4B, ZeRO-2 optimizer state sharding \tabularnewline
\addlinespace[2pt]
MG-GRPO & 2 rd. & $10^{-6}$ & \raggedright $16{\times}8$ per rank, $G=8$, clip $0.2/0.28$, $\lambda=0.3$ for the mediated arm \tabularnewline
\bottomrule
\end{tabular}
\caption{Training configuration by family. Each SFT family covers the four output contracts at three seeds. All runs use four GPUs; SFT runs are full parameter bf16 with per device batch 1 and gradient accumulation to an effective batch of 8. MG-GRPO covers two reward arms over 1{,}500 SoM train prompts, with one clipped policy update per chunk of 16 prompts per rank and no KL term.}
\label{tab:trainconf}
\end{table}

%% file: tables/tableA_seeds.tex
\begin{table}[htbp]
\centering
\small
\setlength{\tabcolsep}{4pt}
\begin{tabular}{llccc}
\toprule
Model & Cond. & seed 1 & seed 2 & seed 3 \\
\midrule
4B & A & .429 & .423 & .417 \\
4B & J & .444 & .426 & .426 \\
4B & JG & .436 & .420 & .419 \\
\addlinespace[2pt]
& $\Delta$ J & $+1.5$ & $+0.3$ & $+0.9$ \\
& $\Delta$ JG & $+0.8$ & $-0.3$ & $+0.2$ \\
\midrule
9B & A & .395 & .411 & .406 \\
9B & J & .404 & .430 & .428 \\
9B & JG & .423 & .426 & .429 \\
\addlinespace[2pt]
& $\Delta$ J & $+0.9$ & $+2.0$ & $+2.1$ \\
& $\Delta$ JG & $+2.8$ & $+1.5$ & $+2.3$ \\
\bottomrule
\end{tabular}
\caption{Per seed SAA behind the means of Table~\ref{tab:multiseed}, with the per seed change from the action only reference in points. Seeds are 20260731 to 20260733. Five of the six model and contract combinations are positive on every seed; the exception is 4B guide first, which is negative on one seed ($-0.3$) and positive on the other two.}
\label{tab:seeds}
\end{table}

%% file: tables/tableA_grpo.tex
\begin{table}[t]
\centering
\small
\setlength{\tabcolsep}{4pt}
\begin{tabular}{lcc}
\toprule
Checkpoint & SAA & BLEU-1 \\
\midrule
Supervised start (JG) & \textbf{.436} & .545 \\
\addlinespace[3pt]
MG-GRPO, round 1 & .426 & .540 \\
MG-GRPO, round 2 & \underline{.433} & .543 \\
\addlinespace[1pt]
Action only, round 1 & .429 & .542 \\
Action only, round 2 & .429 & .545 \\
\bottomrule
\end{tabular}
\caption{Offline reinforcement learning on the Qwen3.5-4B guide first checkpoint. Training reward rises in both arms. The mediated arm ends above the action only control after the second round (underlined), but no checkpoint improves on the supervised starting point.}
\label{tab:grpo}
\end{table}

%% file: webmre.bbl
\begin{thebibliography}{21}
\providecommand{\natexlab}[1]{#1}

\bibitem[{Deng et~al.(2023)Deng, Gu, Zheng, Chen, Stevens, Wang, Sun, and
  Su}]{deng2023mind2web}
Xiang Deng, Yu~Gu, Boyuan Zheng, Shijie Chen, Sam Stevens, Boshi Wang, Huan
  Sun, and Yu~Su. 2023.
\newblock Mind2web: Towards a generalist agent for the web.
\newblock \emph{Advances in Neural Information Processing Systems},
  36:28091--28114.

\bibitem[{Gan et~al.(2026{\natexlab{a}})Gan, Lee, Yin, Liang, He, Wei, Lim,
  Wang, Huang, Zhang, Ni, and Mori}]{gan-etal-2026-multilingual}
Chengguang Gan, Sunbowen Lee, Qingyu Yin, Yunhao Liang, Xinyang He, Hanjun Wei,
  Younghun Lim, Shijian Wang, Hexiang Huang, QingHao Zhang, Shiwen Ni, and
  Tatsunori Mori. 2026{\natexlab{a}}.
\newblock \href {https://doi.org/10.18653/v1/2026.findings-acl.88} {A
  multilingual dataset and empirical validation for the mutual reinforcement
  effect in information extraction}.
\newblock In \emph{Findings of the {A}ssociation for {C}omputational
  {L}inguistics: {ACL} 2026}, pages 1810--1829, San Diego, California, United
  States. Association for Computational Linguistics.

\bibitem[{Gan et~al.(2026{\natexlab{b}})Gan, Wei, Liang, Cai, Zhang, and
  Ni}]{gan2026mag}
Chengguang Gan, Hanjun Wei, Yunhao Liang, Zhixi Cai, Qinghao Zhang, and Shiwen
  Ni. 2026{\natexlab{b}}.
\newblock Mag: A web-agent benchmark and harness for multimodal action and
  guide generation.
\newblock \emph{arXiv preprint arXiv:2607.10079}.

\bibitem[{Koh et~al.(2024)Koh, Lo, Jang, Duvvur, Lim, Huang, Neubig, Zhou,
  Salakhutdinov, and Fried}]{koh2024visualwebarena}
Jing~Yu Koh, Robert Lo, Lawrence Jang, Vikram Duvvur, Ming Lim, Po-Yu Huang,
  Graham Neubig, Shuyan Zhou, Russ Salakhutdinov, and Daniel Fried. 2024.
\newblock Visualwebarena: Evaluating multimodal agents on realistic visual web
  tasks.
\newblock In \emph{Proceedings of the 62nd Annual Meeting of the Association
  for Computational Linguistics (Volume 1: Long Papers)}, pages 881--905.

\bibitem[{Lin(2004)}]{lin2004rouge}
Chin-Yew Lin. 2004.
\newblock Rouge: A package for automatic evaluation of summaries.
\newblock In \emph{Text summarization branches out}, pages 74--81.

\bibitem[{Liu et~al.(2025)Liu, Chen, Li, Qi, Pang, Du, Lee, and
  Lin}]{liu2025understanding}
Zichen Liu, Changyu Chen, Wenjun Li, Penghui Qi, Tianyu Pang, Chao Du, Wee~Sun
  Lee, and Min Lin. 2025.
\newblock Understanding r1-zero-like training: A critical perspective.
\newblock \emph{arXiv preprint arXiv:2503.20783}.

\bibitem[{L{\`u} et~al.(2024)L{\`u}, Kasner, and Reddy}]{lu2024weblinx}
Xing~Han L{\`u}, Zden{\v{e}}k Kasner, and Siva Reddy. 2024.
\newblock Weblinx: Real-world website navigation with multi-turn dialogue.
\newblock \emph{arXiv preprint arXiv:2402.05930}.

\bibitem[{Lu et~al.(2026)Lu, Chai, Guo, Yin, Liu, Wang, Xiao, Ren, Zhao, Liu
  et~al.}]{lu2026ui}
Zhengxi Lu, Yuxiang Chai, Yaxuan Guo, Xi~Yin, Liang Liu, Hao Wang, Han Xiao,
  Shuai Ren, Pengxiang Zhao, Guangyi Liu, et~al. 2026.
\newblock Ui-r1: Enhancing efficient action prediction of gui agents by
  reinforcement learning.
\newblock In \emph{Proceedings of the AAAI Conference on Artificial
  Intelligence}, volume~40, pages 17608--17616.

\bibitem[{Luo et~al.(2025)Luo, Wang, He, Chen, Li, and Xia}]{luo2025gui}
Run Luo, Lu~Wang, Wanwei He, Longze Chen, Jiaming Li, and Xiaobo Xia. 2025.
\newblock Gui-r1: A generalist r1-style vision-language action model for gui
  agents.
\newblock \emph{arXiv preprint arXiv:2504.10458}.

\bibitem[{Nguyen et~al.(2025)Nguyen, Chen, Wang, Wu, Park, Hu, Lyu, Wu, Aponte,
  Xia et~al.}]{nguyen2025gui}
Dang Nguyen, Jian Chen, Yu~Wang, Gang Wu, Namyong Park, Zhengmian Hu, Hanjia
  Lyu, Junda Wu, Ryan Aponte, Yu~Xia, et~al. 2025.
\newblock Gui agents: A survey.
\newblock In \emph{Findings of the Association for Computational Linguistics:
  ACL 2025}, pages 22522--22538.

\bibitem[{Papineni et~al.(2002)Papineni, Roukos, Ward, and
  Zhu}]{papineni2002bleu}
Kishore Papineni, Salim Roukos, Todd Ward, and Wei-Jing Zhu. 2002.
\newblock Bleu: a method for automatic evaluation of machine translation.
\newblock In \emph{Proceedings of the 40th annual meeting of the Association
  for Computational Linguistics}, pages 311--318.

\bibitem[{Qi et~al.(2025)Qi, Liu, Iong, Lai, Sun, Sun, Yang, Yang, Yao, Xu
  et~al.}]{qi2025webrl}
Zehan Qi, Xiao Liu, Iat~Long Iong, Hanyu Lai, Xueqiao Sun, Jiadai Sun, Xinyue
  Yang, Yu~Yang, Shuntian Yao, Wei Xu, et~al. 2025.
\newblock Webrl: Training llm web agents via self-evolving online curriculum
  reinforcement learning.
\newblock In \emph{International Conference on Learning Representations},
  volume 2025, pages 79791--79821.

\bibitem[{Qin et~al.(2025)Qin, Ye, Fang, Wang, Liang, Tian, Zhang, Li, Li,
  Huang et~al.}]{qin2025ui}
Yujia Qin, Yining Ye, Junjie Fang, Haoming Wang, Shihao Liang, Shizuo Tian,
  Junda Zhang, Jiahao Li, Yunxin Li, Shijue Huang, et~al. 2025.
\newblock Ui-tars: Pioneering automated gui interaction with native agents.
\newblock \emph{arXiv preprint arXiv:2501.12326}.

\bibitem[{Shao et~al.(2024)Shao, Wang, Zhu, Xu, Song, Bi, Zhang, Zhang, Li, Wu
  et~al.}]{shao2024deepseekmath}
Zhihong Shao, Peiyi Wang, Qihao Zhu, Runxin Xu, Junxiao Song, Xiao Bi, Haowei
  Zhang, Mingchuan Zhang, YK~Li, Yang Wu, et~al. 2024.
\newblock Deepseekmath: Pushing the limits of mathematical reasoning in open
  language models.
\newblock \emph{arXiv preprint arXiv:2402.03300}.

\bibitem[{Team(2026)}]{qwen35blog}
Qwen Team. 2026.
\newblock \href {https://qwen.ai/blog?id=qwen3.5} {Qwen3.5: Accelerating
  productivity with native multimodal agents}.

\bibitem[{Wei et~al.(2025)Wei, Yao, Liu, Zhang, Lu, Qiu, Yu, Xu, Zhang, Yin
  et~al.}]{wei2025webagent}
Zhepei Wei, Wenlin Yao, Yao Liu, Weizhi Zhang, Qin Lu, Liang Qiu, Changlong Yu,
  Puyang Xu, Chao Zhang, Bing Yin, et~al. 2025.
\newblock Webagent-r1: Training web agents via end-to-end multi-turn
  reinforcement learning.
\newblock In \emph{Proceedings of the 2025 Conference on Empirical Methods in
  Natural Language Processing}, pages 7920--7939.

\bibitem[{Xu et~al.(2024{\natexlab{a}})Xu, Lu, Shen, Wang, Wang, Mao, Xiong,
  and Yu}]{xu2024agenttrek}
Yiheng Xu, Dunjie Lu, Zhennan Shen, Junli Wang, Zekun Wang, Yuchen Mao, Caiming
  Xiong, and Tao Yu. 2024{\natexlab{a}}.
\newblock Agenttrek: Agent trajectory synthesis via guiding replay with web
  tutorials.
\newblock \emph{arXiv preprint arXiv:2412.09605}.

\bibitem[{Xu et~al.(2024{\natexlab{b}})Xu, Wang, Wang, Lu, Xie, Saha, Sahoo,
  Yu, and Xiong}]{xu2024aguvis}
Yiheng Xu, Zekun Wang, Junli Wang, Dunjie Lu, Tianbao Xie, Amrita Saha, Doyen
  Sahoo, Tao Yu, and Caiming Xiong. 2024{\natexlab{b}}.
\newblock Aguvis: Unified pure vision agents for autonomous gui interaction.
\newblock \emph{arXiv preprint arXiv:2412.04454}.

\bibitem[{Yu et~al.(2025)Yu, Zhang, Zhu, Yuan, Zuo, Yue, Dai, Fan, Liu, liu,
  Liu, Liu, Lin, Lin, Ma, Sheng, Tong, Zhang, Zhang, Zhang, Zhang, Zhu, Zhu,
  Chen, Chen, Wang, Yu, Song, Wei, Zhou, Liu, Ma, Zhang, Yan, Wu, and
  Wang}]{NEURIPS2025_a4277440}
Qiying Yu, Zheng Zhang, Ruofei Zhu, Yufeng Yuan, Xiaochen Zuo, Yu~Yue, Weinan
  Dai, Tiantian Fan, Gaohong Liu, juncai liu, LingJun Liu, Xin Liu, Haibin Lin,
  Zhiqi Lin, Bole Ma, Guangming Sheng, Yuxuan Tong, Chi Zhang, Mofan Zhang, and
  17 others. 2025.
\newblock \href
  {https://proceedings.neurips.cc/paper_files/paper/2025/file/a4277440d50f1f15d2cb4c14f7e0c0d2-Paper-Conference.pdf}
  {Dapo: An open-source llm reinforcement learning system at scale}.
\newblock In \emph{Advances in Neural Information Processing Systems},
  volume~38, pages 113222--113244. Curran Associates, Inc.

\bibitem[{Zhang et~al.(2024)Zhang, Wu, Yihua, Liao, Xu, Xiao, Wei, and
  Tang}]{zhang2024android}
Jiwen Zhang, Jihao Wu, Teng Yihua, Minghui Liao, Nuo Xu, Xiao Xiao, Zhongyu
  Wei, and Duyu Tang. 2024.
\newblock Android in the zoo: Chain-of-action-thought for gui agents.
\newblock In \emph{Findings of the Association for Computational Linguistics:
  EMNLP 2024}, pages 12016--12031.

\bibitem[{Zhou et~al.(2024)Zhou, Xu, Zhu, Zhou, Lo, Sridhar, Cheng, Ou, Bisk,
  Fried et~al.}]{zhou2024webarena}
Shuyan Zhou, Frank~F Xu, Hao Zhu, Xuhui Zhou, Robert Lo, Abishek Sridhar,
  Xianyi Cheng, Tianyue Ou, Yonatan Bisk, Daniel Fried, et~al. 2024.
\newblock Webarena: A realistic web environment for building autonomous agents.
\newblock In \emph{International Conference on Learning Representations},
  volume 2024, pages 15585--15606.

\end{thebibliography}
